\documentclass[twocolumn, switch]{article} 

\usepackage{preprint}

\usepackage{amsmath, amsthm, amssymb, amsfonts}
\usepackage[ruled,vlined,norelsize,\languagename]{algorithm2e}

\usepackage[numbers,square]{natbib}
\usepackage{natbib}

\usepackage[utf8]{inputenc}	
\usepackage[T1]{fontenc}	
\usepackage{xcolor}		
\usepackage[colorlinks = true,
            linkcolor = purple,
            urlcolor  = blue,
            citecolor = cyan,
            anchorcolor = black]{hyperref}	
\usepackage{booktabs} 		
\usepackage{nicefrac}		
\usepackage{microtype}		
\usepackage{lineno}		
\usepackage{float}			
\usepackage{graphicx}
\usepackage[hang,small,bf,tight]{subfigure}
\usepackage[capitalise]{cleveref}
\usepackage{tabularx}
\usepackage{enumitem}
\setlist{topsep=0pt, leftmargin=*}

\usepackage{lipsum}		

\usepackage{newfloat}
\DeclareFloatingEnvironment[name={Supplementary Figure}]{suppfigure}
\usepackage{sidecap}
\sidecaptionvpos{figure}{c}

\usepackage{titlesec}
\titlespacing\section{0pt}{12pt plus 3pt minus 3pt}{1pt plus 1pt minus 1pt}
\titlespacing\subsection{0pt}{10pt plus 3pt minus 3pt}{1pt plus 1pt minus 1pt}
\titlespacing\subsubsection{0pt}{8pt plus 3pt minus 3pt}{1pt plus 1pt minus 1pt}

\usepackage{tikz,xcolor,hyperref}

\definecolor{lime}{HTML}{A6CE39}

\title{Modelling the Recommender Alignment Problem}

\usepackage{authblk}

\author[1]{Francisco Carvalho}

\affil[1]{Departmento de Engenharia Informática, Instituto Superior Técnico}

\begin{document}

\twocolumn[ 
  \begin{@twocolumnfalse} 
  
\maketitle

\begin{abstract}
\noindent Recommender systems (RS) mediate human experience online. Most RS act to optimize metrics that are imperfectly aligned with the best-interest of users but are easy to measure, like ad-clicks and user engagement. This has resulted in a host of hard-to-measure side-effects: political polarisation \cite{benkler_network_2018}, addiction \cite{hasan_excessive_2018} \cite{andreassen_online_2015-1}, fake news \cite{stocker_how_2020}. RS design faces a \textit{recommender alignment problem}: that of aligning recommendations with the goals of users, system designers, and society as a whole. But how do we test and compare potential solutions to align RS? Their massive scale makes them costly and risky to test in deployment. We synthesized a simple abstract modelling framework to guide future work. 

To illustrate it, we construct a toy experiment where we ask: "How can we evaluate the consequences of using user retention as a reward function?" To answer the question, we learn recommender policies that optimize reward functions by controlling graph dynamics on a toy environment. Based on the effects that trained recommenders have on their environment, we conclude that engagement maximizers generally lead to worse outcomes than aligned recommenders but not always. After learning, we examine competition between RS as a potential solution to RS alignment. We find that it generally makes our toy-society better-off than it would be under the absence of recommendation or engagement maximizers. 

In this work, we aimed for a broad scope, touching superficially on many different points to shed light on how an end-to-end study of reward functions for recommender systems might be done. Recommender alignment is a pressing and important problem. Attempted solutions are sure to have far-reaching impacts. Here, we take a first step in developing methods to evaluating and comparing solutions with respect to their impacts on society.

\end{abstract}
\vspace{0.35cm}

  \end{@twocolumnfalse} 
] 



\section{Introduction}
\label{sec:introduction}
Recommender systems (RS) are software systems that assist users in interacting with large spaces of items, usually by presenting them with smaller personalized sets based on information such as past user behavior, user attributes, and features of the underlying items. User experience on social media, content platforms, and online stores is largely determined by RS.  

Most recommender systems optimize metrics that are easy to measure and improve, like number of clicks, time spent, or number of daily active users. 
They are selected to do this by powerful optimization processes involving thousands of engineers and a significant fraction of global computing power.
Goodhart's law \cite{goodhart_problems_1981} states that "when a metric becomes a goal, it ceases to be a good metric". In fact, choosing metrics that are imperfectly aligned with the best-interest of users has resulted in a host of hard-to-measure side-effects like political polarisation \cite{benkler_network_2018}, addiction \cite{hasan_excessive_2018} \cite{andreassen_online_2015-1}, fake news \cite{stocker_how_2020}, fairness \cite{barocas-hardt-narayanan} and diversity\cite{castells_novelty_2015-1} concerns.


Recommender system design faces a \textit{recommender alignment problem}: that of aligning recommendations with the goals of users, system designers, and society as a whole. We conceive it in analogy to the \textit{value alignment problem} \cite{hadfield-menell_incomplete_2019}: that of ensuring that an AI system's behavior aligns with the values of its principal. 

\begin{figure}[h]
\centering
\includegraphics[width=0.9\linewidth]{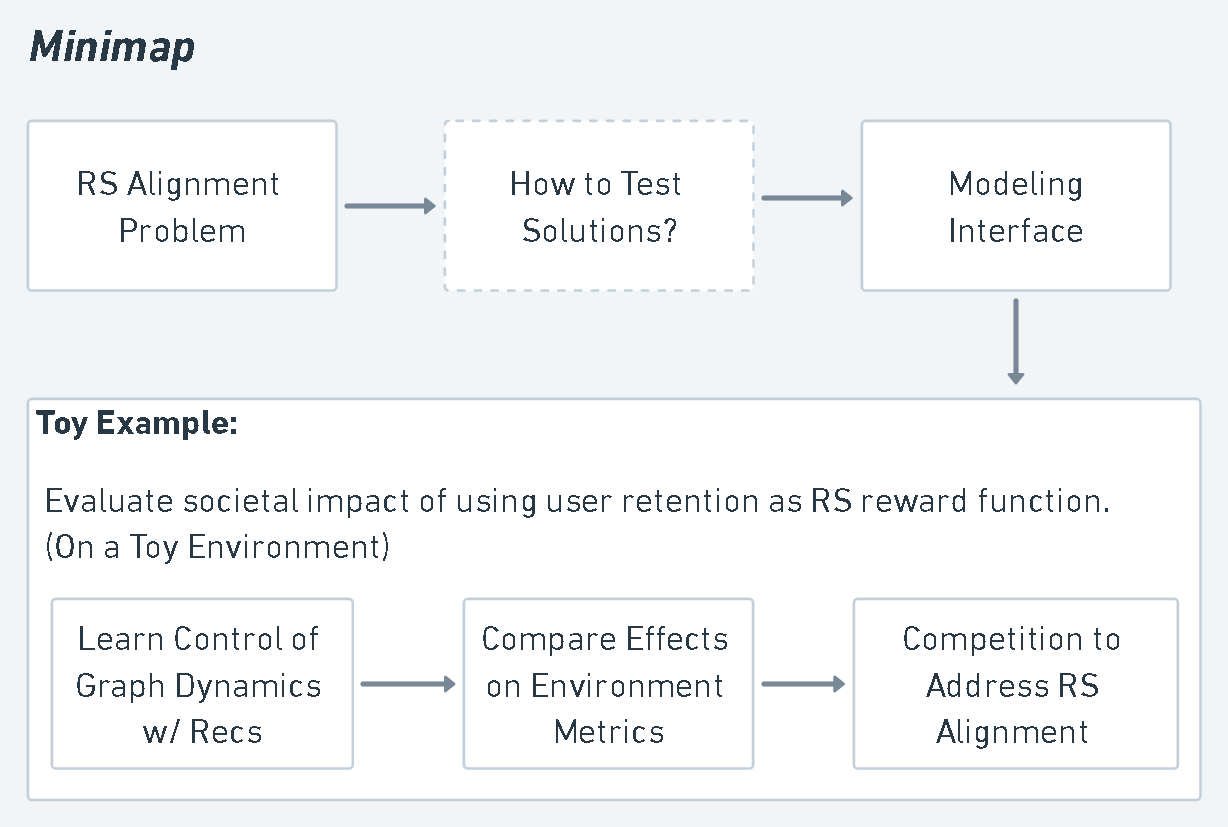}
\caption{\textbf{How our contributions relate}. Potential solutions to the RS alignment problem will need to be evaluated with modeling and simulation. We construct a modeling interface based on qualitative literature on RS impacts. To illustrate its use, we run an end-to-end experiment to evaluate the effects of user retention as a reward function. We learn recommender policies aligned and misaligned with society in a toy environment and analyse relevant metrics. We further examine competition as a potential solution to alignment in our environment.}
\label{fig:minimap}
\end{figure}

A single reward function that would work for everyone forever is unlikely to exist. Thus a reward function for RS should be adaptive and error-correcting, drawing adjustments in a bottom-up fashion subject to users themselves. For example: recommender systems are currently unopposed in their respective networks. Would enabling competition between RS result in a sufficiently adaptive system?

But how do we test and compare potential solutions to align RS? How can we learn about their impacts on society and emergent effects? The massive scale of these systems makes them costly (due to the amount of resources involved), risky (due to the number of users they impact) and arguably unethical to test in deployment. Modelling and simulation approaches are used under similar constraints to study social dynamics so we consider them suitable for our case as well.

We present a modelling interface: a minimal set of requirements for models, derived from the literature around the societal effects of RS. The interaction between RS and society is a complex, multifaceted topic. What common properties of recommender systems are essential to study alignment, and which ones are domain-specific factors that should be left for implementation? We could find no prior efforts to base ourselves on, so we hope our interface can be relied on for future attempts to evaluate alignment techniques.

Our contributions lie in addressing the following questions. Our modeling interface is concerned with 1, while 2, 3, and 4 concern our toy-experiment.

\begin{enumerate}
\item \textbf{Modeling Interface}: How can models focus on dynamics relevant to RS alignment while abstracting over domain-specific factors?
\item \textbf{Toy environment}: How can we evaluate the societal implications of having user retention as the reward function for a recommender?
\item \textbf{Learning}: Can we learn policies that control graph dynamics to optimize arbitrary rewards?
\item \textbf{Competition}: Will competition between RS lead to better outcomes for society than recommender monopolies?
\end{enumerate}


\section{Modelling Interface}
\label{sec:modeling}
How can models focus on dynamics relevant to RS alignment, while abstracting over domain-specific factors? Finding no prior efforts to base ourselves on, we present a novel modeling interface intended to inform future modeling work on recommender alignment techniques. 

This minimal set of requirements was derived from the literature around the societal effects of RS. We established a minimal set of relevant entities from prior views on RS as multi-stakeholder environments. \cite{milano_recommender_2020} To abstract interactions between entities, we relied on an existing taxonomy of human interactions with Intelligent Software Agents (ISA). \cite{burr_analysis_2018} 

\begin{figure}[h]
\centering
\includegraphics[width=0.9\linewidth]{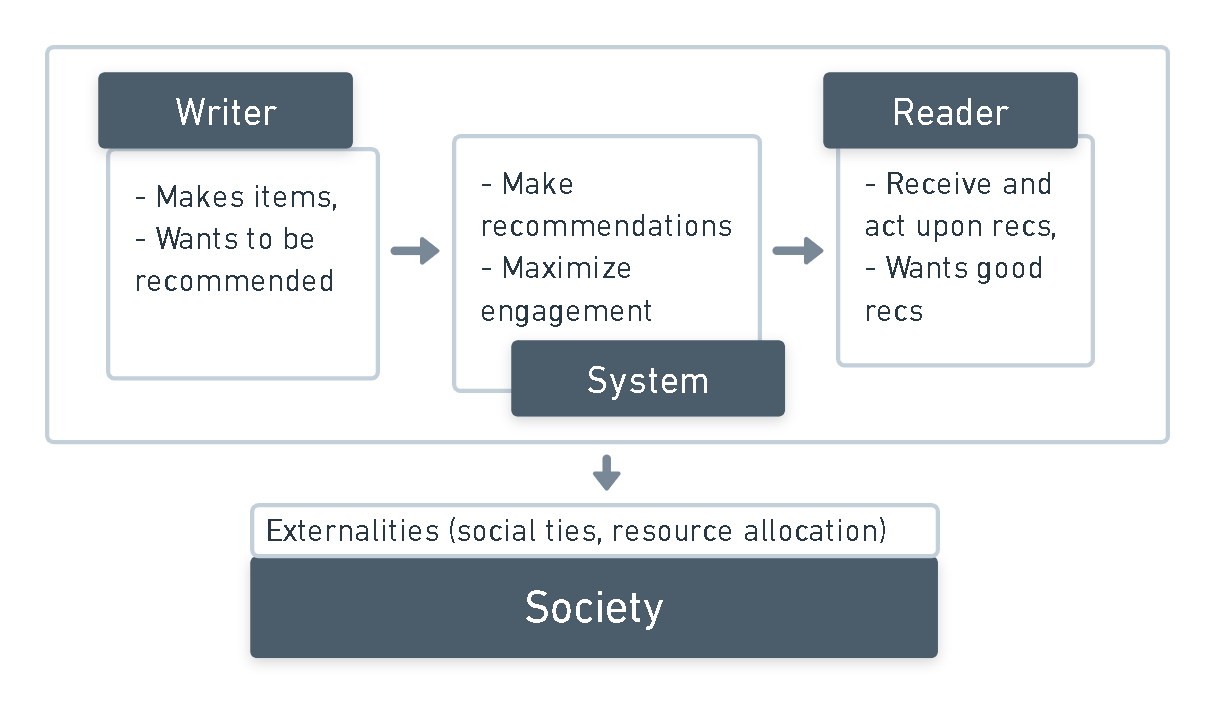}
\caption{Recommender systems as peer-to-peer multi-stakeholder environments. \cite{milano_recommender_2020} In p2p environments like Twitter, for example, users are both content producers and consumers with interests in both being recommended to other users, and to receive recommendations that advance their goals (entertainment, well-being, learning, etc). RS mediate information transfer between users and their non-local environment. The operators of RS have an interest in keeping users engaged, irrespective of whether that is in their best long-term interest or not. Finally, externalities from these interactions are expressed in society, as people take action as a result of things they learned, opinions they formed, or relationships they started online.}
\label{fig:p2p_stakeholders}
\end{figure}

\begin{table}[htbp]
\caption{Modeling interface requirements. } \label{tab:translation}
\resizebox{0.55\linewidth}{!}{\begin{tabular}{l}
\toprule
\textbf{Components} \\
1a) \textbar \space Environment\\
1b) \textbar \space Users \\
1c) \textbar \space Recommender\\
\textbf{Interaction} \\
2a) \textbar \space User-RS interaction \\
2b) \textbar \space User-Environment interaction\\
\textbf{Utility functions} \\
3a) \textbar \space RS utility \\
3b) \textbar \space User utility \\
3c) \textbar \space Social utility \\
\bottomrule
\end{tabular}}
\end{table}

\textbf{Entities: an Environment, Users with local information and Recommenders with system-level information.}
An environment should contain the users which act in it and provide observations for users and recommenders. 
The task of RS is to parse large spaces of items and present personalized subsets to users, who are unable to observe the whole space. It is sensible then to model users with local information and RS with access to more information, or even global information.

\bigskip

\textbf{User-RS interaction: recommendations}. 
RS mediate the relationship between users and their wider environment. They do this by providing information about it in the form of recommendations. Recommendations can be items of content or other users to interact with (as it happens with friend suggestions in social networks). 

\bigskip

\textbf{User-Environment interaction}.
The dynamical system of society. People exist and pursue their goals in the world. On social media platforms, experience consists of interacting with other users either directly or by consuming and producing content. The RS largely mediates this experience. Although an increasing part of people's lives and the economy is conducted online under mediation, the offline world represents a source and sink of externalities away from the reach of RS. As users interact with RS, they change the way they act in the world, and the environment is changed in turn. It's by measuring this change that we can evaluate the impacts of RS.

\bigskip

\textbf{Utility functions for recommenders, users, and society}. 
RS are selected by their operators to improve some metric, either by human selection or by reinforcement learning algorithms. In the case of many social media platforms this metric is a proxy that benefits whoever is in charge of the RS - usually related to click-through rate on advertisements or time spent by users on the platform.
Defining utility for even a single individual is challenging under the full complexity of the real world. There has been work on optimizing for self-actualization or self-reported well-being, but ultimately, realistic notions of individual utility must include regular feedback and redefinition. 
A notion of utility for society is even more challenging to achieve. People often have incompatible goals and values that must be traded-off even when making decisions with complete information. Nonetheless, we must assume these can be approximated and represented in models.

\begin{figure}[h]
\centering
\includegraphics[width=0.8\linewidth]{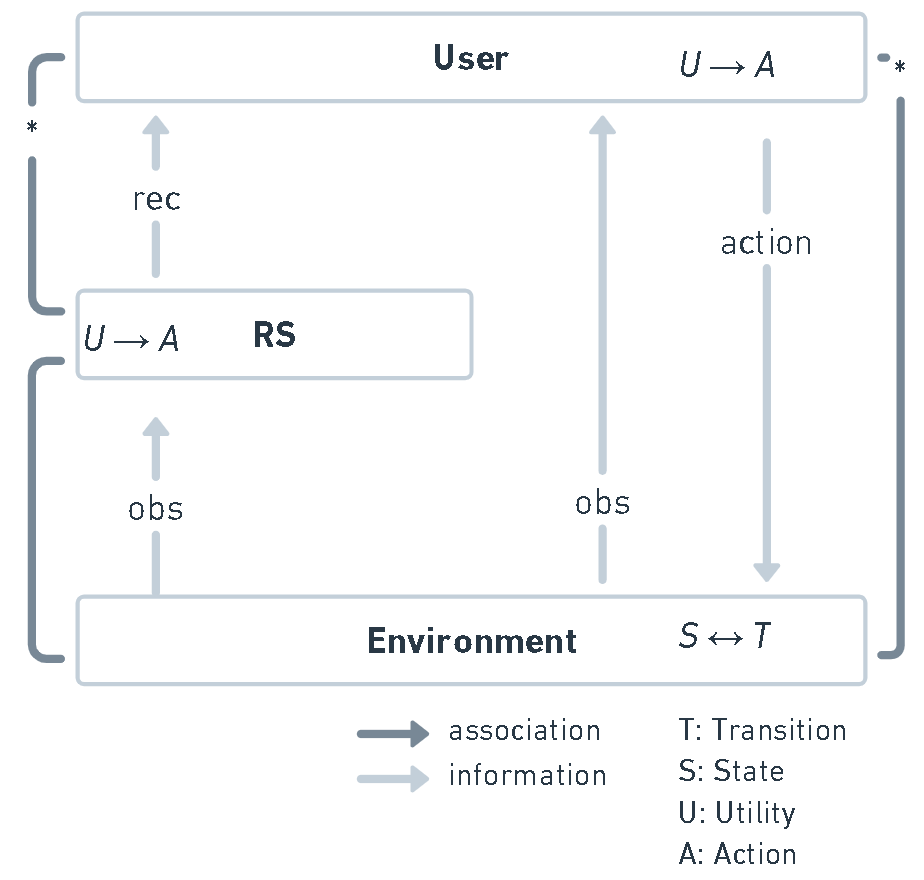}
\caption{UML diagram of the relevant entities in our modeling interface: \textit{Users}, \textit{RS}, and the \textit{Environment}. User has as any-to-one relationship with both RS and the environment. Both User and RS have \textit{utility functions} that determine their actions. The environment has \textit{State} and a \textit{Transition function} $T: \{S,A\} \longrightarrow S$. Information about state is transmitted to User and RS via observations of different portions of the environment. RS make \textit{recommendations} to Users providing extra information about State, Users \textit{act} on the Environment.   }
\label{fig:uml_small}
\end{figure}

\section{Experiment: Environment}
\label{sec:environment}
In this section, we define a concrete model that implements the interface laid out in \Cref{sec:modeling}. To do this, we extend an existing framework of networked evolutionary game theory \cite{santos_cooperation_2006} where agents play dilemmas of cooperation, periodically rewiring social ties. Our main extension to the original model is allowing arbitrary rewiring policies to be used instead of the original default policy, thus allowing us to delegate the choice of new neighbors to a recommender system. Finally, we obtain baselines for the behavior of the system. To get a comprehensive picture, we simulate the environment as mediated by different fixed rewiring heuristics using the information in strategies and node degrees.

\subsection{Definition}

\begin{figure}[h]
\centering
\includegraphics[width=0.9\linewidth]{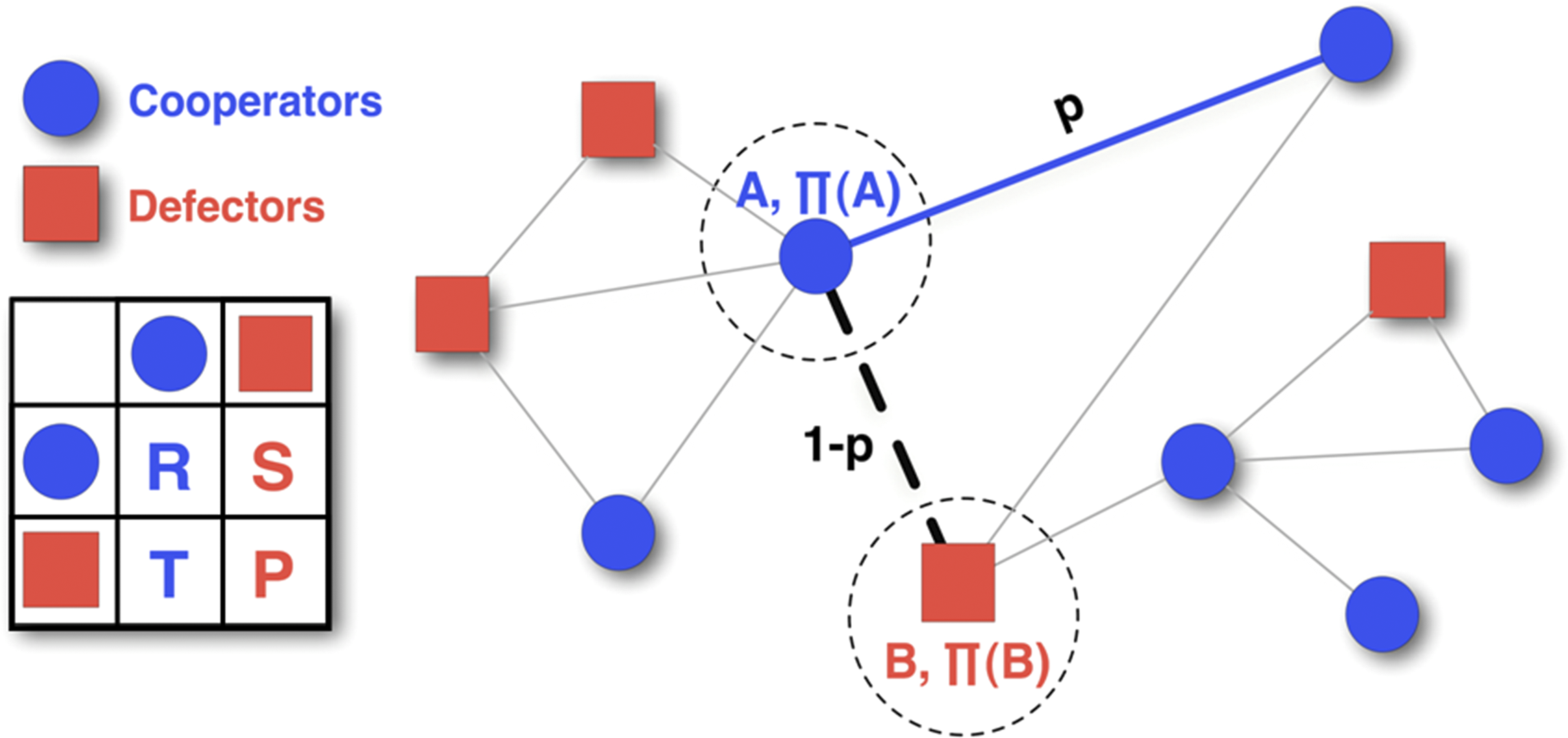}
\caption{Our base model. Users play dilemmas of cooperation where they have two possible strategies: cooperate or defect. They receive payoff from playing other agents, which determines their fitness. A user A will also try to rewire their edge to B if B is a Defector. Whether a rewire succeeds depends on the difference of fitness of the nodes involved. Adapted from \cite{santos_cooperation_2006}. }
\label{fig:neighborhood}
\end{figure}

There are two types of individuals: cooperators and defectors, we call their strategies $C$ and $D$ respectively. They engage in social dilemmas of cooperation - specifically 2-player symmetric games - where players can either cooperate or defect when interacting. (see \Cref{fig:neighborhood})
Individuals only interact with their neighbors on the network. 
Game strategies evolve in the population as users compare fitness and copy their neighbors' strategies. 
Individuals can also rewire their social ties if unsatisfied with their neighbors.

\paragraph{Network} Users are connected to one another according to the edges of a network graph $G={V,E}$ where $V$ is the set of nodes and $E$ is the set of edges. $G$ is always initialized as a uniform random graph with average node degree $k$ and its topology is allowed to evolve.

\paragraph{Games} Agents interact by playing social dilemmas: symmetric, 2-player, 2x2 matrix games. (as seen in \Cref{fig:neighborhood}) 
We normalize the difference between mutual cooperation ($R$) and mutual defection ($P$) to 1, making $R = 1$ and $P = 0$, respectively. 
As a consequence, games can be parameterized by two scalars: payoff $T$ (temptation to cheat), which satisfies $0 \leq T \leq 2$ and payoff $S$ (disadvantage of being cheated) satisfies $-1 \leq S \leq 1$. 
In this paper, we will focus on the Prisoner's Dilemma : $T=2$, $S=-1$ as it is the hardest game to solve. We have however, explored the full space of parameters for some experiments.

\paragraph{Evolution} The strategy of a node x evolves through imitation of a neighbor y. A node updates its strategy according to a Fermi update probability $p$ (\cref{eqn:fermi}) based on the difference between the fitness of each player ($f_A$ and $f_B$). \cite{traulsen_stochastic_2006}  Fitness corresponds to the cumulative payoffs of a node, resulting from the sum of payoffs from playing each of one’s neighbors.

\begin{equation}
\label{eqn:fermi}
p = \frac{1}{1+e^{-\beta(f_B - f_A)}}
\end{equation}

\paragraph{Rewiring} Given an edge between A and B, we say A is satisfied with the link if B is a cooperator, being dissatisfied otherwise. 
If A is satisfied, they will keep the link. If dissatisfied, A will compete with B to rewire the link. (\Cref{fig:neighborhood}) 
The action taken is contingent on the fitness $\Pi(A)$ and $\Pi(B)$ of A and B respectively. 
A redirects the link to a new neighbor given by its rewiring strategy with probability p given by \cref{eqn:fermi}.
With probability $1 - p$, A either stays linked to B - if A is a cooperator - or B rewires its link with A to one of A's neighbors. We call this rewiring a \textit{structural update}.

\paragraph{Time-scale} Strategy evolution and structural evolution can occur at different time-scales, $T_a$ and $T_e$ respectively (if $T_a = 2*T_E$, strategy updates occur twice as often as structural ones).
The ratio $W = T_e/T_a$, leads to different outcomes for cooperation. In realistic situations, the two time-scales should be of comparable magnitude. $W$ serves as a measure of agents' inertia to react to their conditions: large values of $W$ reflect populations where individuals - on average - react promptly to adverse ties, whereas smaller values reflect some inertia for rewiring social ties.

\begin{table}[htbp]
\centering
\caption{Translation table between problem and model space. The third column is a concrete example about the real world.}\label{tab:translation}
\resizebox{0.9\linewidth}{!}{\begin{tabular}{ll}
\toprule
\textbf{Problem space} &  \textbf{Model space}\\
\midrule
Recommenders & Rewiring policies\\
Users & Nodes \\
Environment & Nodes in Network\\
User-environment interaction & 2x2 game with neighbor\\
Item recommendation & Neighbor recommendation\\
User utility & Game payoff\\
Societal utility & Cooperator ratio\\
User engagement with RS & Number of rewires\\
\bottomrule
\end{tabular}}
\end{table}
\subsection{Simulation algorithm}
\begin{algorithm}[ht]
\DontPrintSemicolon
\BlankLine
\While{$t < timeLimit$}{
    $x \longleftarrow randomSample(V)$\;
    $y \longleftarrow randomNeighbor(x)$\;
    $P_x, P_y \longleftarrow cumulativePayoff(x), cumulativePayoff(y)$\;
    $p \longleftarrow fermi(P_x - P_y, beta)$\;
    \BlankLine
    \If{$random(0,1) < (1+W)^{-1}$}{
        \If{$random(0,1) < p$}{
            $Strat_x \longleftarrow  Strat_y$\;
        }
    }
    \Else{
        \If{$Strat_y == D \, and \,   Strat_x == C$}{
            \If{$random(0,1) < p$}{
                $z \longleftarrow rewireStrat_x(x,y)$\;
                $doRewire(G, x, y, z)$\;
            }
        }
        \If{$Strat_y == D \, and \,  Strat_x == D$}{
            \If{$random(0,1) < p$}{
                $z \longleftarrow rewireStrat_x(x,y)$\;
                $doRewire(G, x, y, z)$\;
            }
            \Else{
                $z \longleftarrow rewireStrat_y(y,x)$\;
                $doRewire(G, y, x, z)$\;
            }
        }
    }
}
\label{alg:simulation}
\caption{Simulation step algorithm}
\end{algorithm}

The pseudo code for the update process in our simulations is described in \cref{alg:simulation}. 
$fermi(A,B, beta)$ is the function that calculates \cref{eqn:fermi} given A, B, and temperature term $beta$.
$cumulativePayoff(x)$ returns the sum of payoffs a node $x$ gets after playing a game with each of its neighbors. 
$W$ is the ratio between the timescales of structural evolution and strategy evolution: $W = T_e/T_a$.
$Strat$ is a vector of strategies (taking values in $\{C,D\}$) and $rewireStrat$ is a vector of rewiring strategies, each of these has a length $\#V$. A rewiring strategy is a function $R : \{x, y\} -> z$ that provides a recommended node $z$ given a focused node $x$ and the neighbor being rewired $y$. 
$doRewire(G,x,y,z)$ deletes the edge $(x,y)$ from G and adds edge $(x,z)$.

\subsection{Heuristics and Baselines}

\begin{figure}[h]
\centering
\includegraphics[width=1\linewidth]{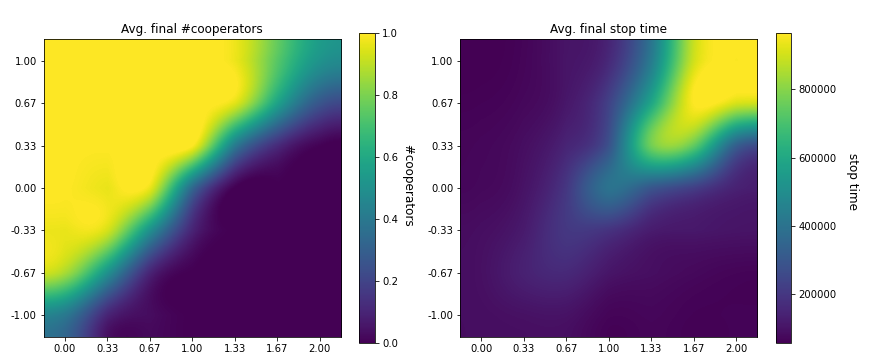}
\caption{Evolution of cooperation in a uniform random network in the absence of rewiring (W=0). On the left: average final \textbf{number of cooperators}. On the right: final \textbf{stop time}. Both plotted as a function of game-parameters: S, the disadvantage of a cooperator being defected (when $S < 0$), and T, the temptation to defect on a cooperator (when $T > 1$). Absent any of these threats ($S \geq 0$ and $T \leq 1$; upper-left quadrant) cooperators trivially dominate. The lower-left quadrant ($S < 0$ and $T \leq 1$) corresponds to the Stag-Hunt dilemma, by definition. The lower triangle in the upper-right quadrant ($S \geq 0$, $T > 1$) corresponds to the Snowdrift game, also by definition. The lower-right quadrant ($S < 0$ and $T > 1$) corresponds to the Prisoner’s Dilemma domain (PD). \cite{santos_evolutionary_2006} 
}
\label{fig:norewire}
\end{figure}

We focus the rest of our simulations on the Prisoner's Dilemma. We evaluate a small set of recommender heuristics: BAD (always recommends random defectors), RANDOM (always recommends random nodes), GOOD (always recommends random cooperators), NO\_MED (local heuristic), and FAIR (recommend random cooperators to cooperators and defectors to defectors). We observe an ordering by speed of convergence towards cooperation.

\paragraph{Local heuristic} \label{par:local} 
The local heuristic corresponds to the default rewiring policy defined in \cite{santos_cooperation_2006}. Given an edge $(A,B)$, node $A$ rewires to a random neighbor of node $B$. 
The intuition behind this reasoning is that simple agents, being rational individuals with partial information, are more likely to interact with nearby agents\cite{kossinets_empirical_2006}.  Moreover, selecting a neighbour of an inconvenient partner is also a good choice, since this partner also tries to establish links with cooperators, making it more likely that the rewiring results in a tie to a cooperator.

\begin{figure}[htbp]
	\centering
	\subfigure[Final fraction of cooperators. The order of convergence speed can be more clearly observed.]{\label{fig:coop_freq_all_meds} 		\includegraphics[width=0.48\linewidth]{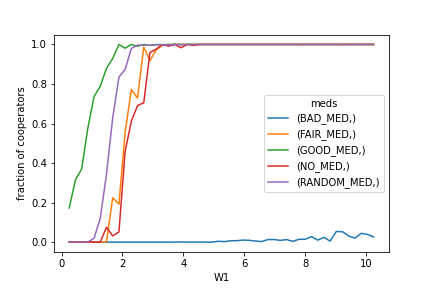}}
	\subfigure[Number of rewires per rewire opportunity. (To normalize with respect to $W$)]{\label{fig:rewires_all_meds}
		\includegraphics[width=0.48\linewidth]{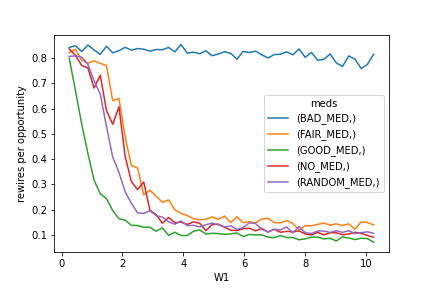}}
	\subfigure[Heterogeneity. NO MED clearly leads to more heterogeneous networks. We expect this to be due to its local focus, rather than global recommendations.]{\label{fig:heterogeneity_all_meds} 		\includegraphics[width=0.48\linewidth]{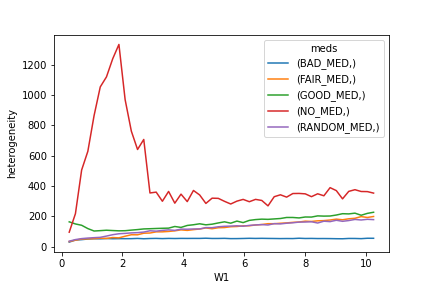}}
	\subfigure[Max degree, another metric of heterogeneity. We can see BAD leads to the absolute lowest heterogeneity.]{\label{fig:k_max_all_meds} 		\includegraphics[width=0.48\linewidth]{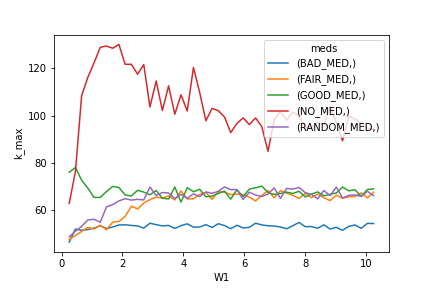}}
	\caption{Comparison of different recommender heuristics. Co-evolution of cooperation and structure in the Prisoner's Dilemma as a function of W. }
	\label{fig:line_over_w}
\end{figure}

In terms of rewiring, BAD leads to the highest rate, while GOOD produces the lowest, with the remaining mediators being comparable (\cref{fig:line_over_w}). One of the reasons is that agents will only want recommendations if the neighbor they're rewiring is a Defector, so a population with more Defectors will want more rewires, while a population that converges to full Cooperators will cease to want new neighbors. 

This means we can expect that recommending Cooperators will drive convergence to cooperation, and that recommending defectors will drive high rewiring numbers until convergence to full defection. From our experiments in the full space of game parameters (not pictured), we observe the Prisoner's Dilemma is the hardest setting for each of our heuristics and that there is no overwhelming convergence to cooperation in any of them when $W=1$.
\newline
Unexpectedly, we observed FAIR does worse than RANDOM with respect to convergence towards cooperation. We believe this is because defectors and cooperators become segregated making it more difficult for defectors to convert. This hasn't been tested but we would start by computing modularity-based community finding and measuring the average strategies in each community. 
\newline
All the heuristics we used were global in scope, having a homogenizing effect, thus NO\_MED generally led to higher heterogeneity than any of them. 
These initial baselines were obtained from a set of 5 heuristics. The following chapters include a larger set of heuristics based on their strategies and degree information. 
For the rest of the experiment, we'll consider a 3-by-3 space of heuristics plus the null policy (no rewiring) and the local heuristic defined in \cref{par:local}.
\begin{table}[htbp]
\centering
\caption{Possible heuristics as a combination of degree and strategy. The Cartesian product between choosing the lowest, random, and highest degree nodes - and choosing only defectors, any strategy, or only cooperators. }\label{tab:translation}
\resizebox{0.6\linewidth}{!}{\begin{tabular}{cccc}
\toprule
\textbf{strat x degree} &  \textbf{Low} &  \textbf{Random} &  \textbf{High}\\
\midrule
\textbf{Defectors only} & X & X & X\\
\textbf{Random} & X & X & X\\
\textbf{Cooperators only} & X & X & X\\
\bottomrule
\end{tabular}}
\end{table}

\section{Experiment: Learning to Control Graph Dynamics with Recommendations}
\label{sec:learning}
Our goal in this section is to use reinforcement learning to learn two kinds of recommender policies: \textit{aligned} policies that maximize the final \textbf{number of cooperators}, and \textit{engagement-maximizer} policies that maximize the \textbf{number of rewires} that take place over runs. We want to learn them so that we can compare their behavior and effects on their environment. Specifically with regards to cooperation, \#rewires, and network topology. 

\subsection{Training Architecture}

\begin{figure}[h]
\centering
\includegraphics[width=1\linewidth]{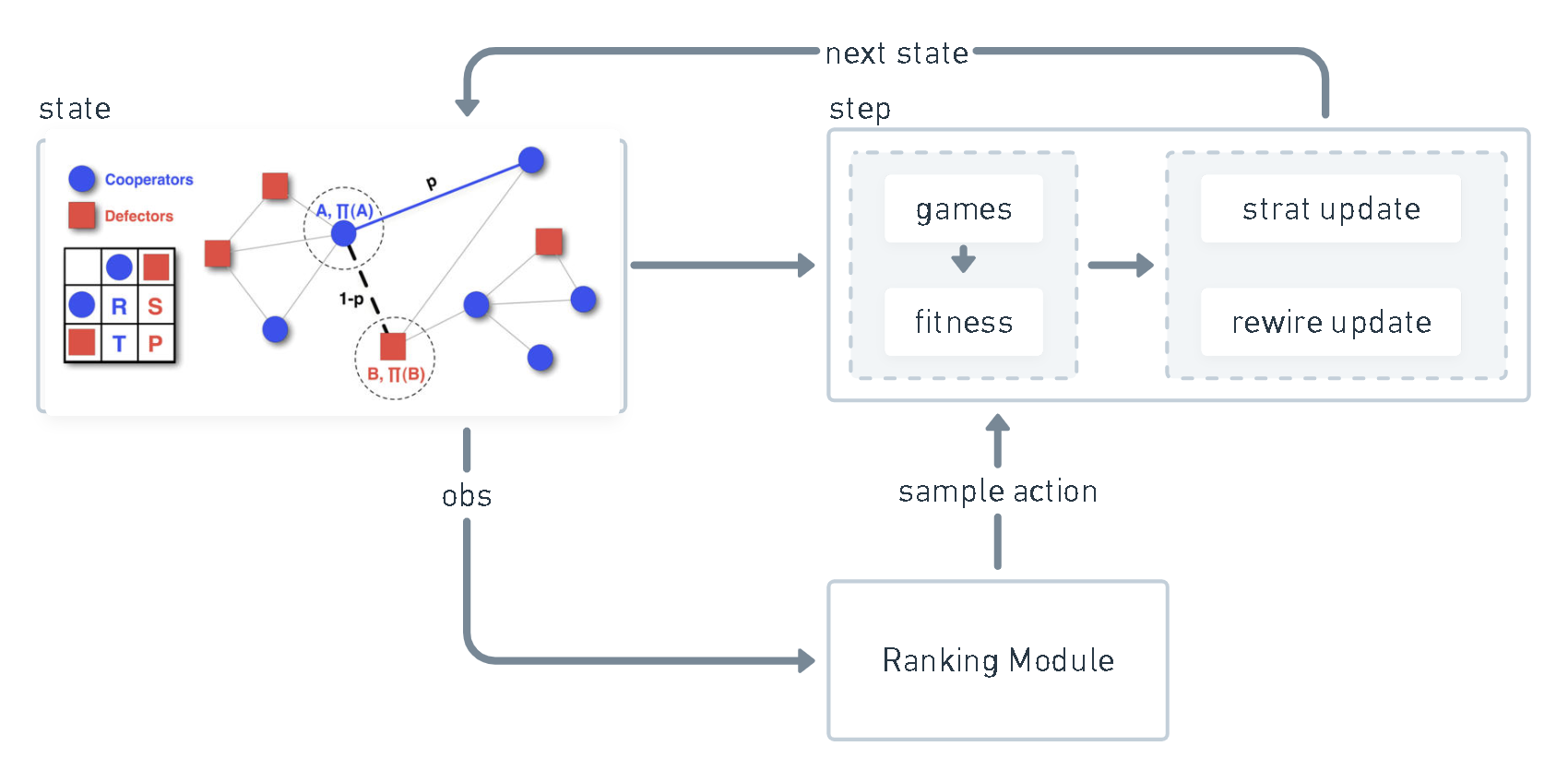}
\caption{Structure of our training loop. Observations at each time-step $t$ contain a focused node $x$, the graph adjacency matrix, and node features. The action consists of selecting a node $z$ to which $x$ will rewire one of its edges. $x$ then goes through a strategy update. $W=1$; $T,S = 2,-1$.}
\label{fig:rl_env}
\end{figure}

\begin{table}[htbp]
\centering
\caption{Environment configurations. $N$ is the number of nodes in the graph, $\beta$ is the temperature parameter for the fermi \cref{eqn:fermi} expression, $k$ is the average node degree. The time limit is the number of steps the environment can take before a simulation is ended (Simulations usually finish early). Environment steps equal number of strategy plus rewire updates. }\label{tab:translation}
\resizebox{0.5\linewidth}{!}{\begin{tabular}{llll}
\toprule
\textbf{N} &  $\beta$ &  \textbf{k} &  \textbf{time limit}\\
\midrule
\textbf{10} & 0.1 & 4 & 1000\\
\textbf{30} & 0.05 & 8 & 3000\\
\textbf{100} & 0.005 & 28 & 10000\\
\textbf{500} & 0.005 & 30 & 30000\\
\bottomrule
\end{tabular}}
\end{table}

\begin{figure}[h]
\centering
\includegraphics[width=1\linewidth]{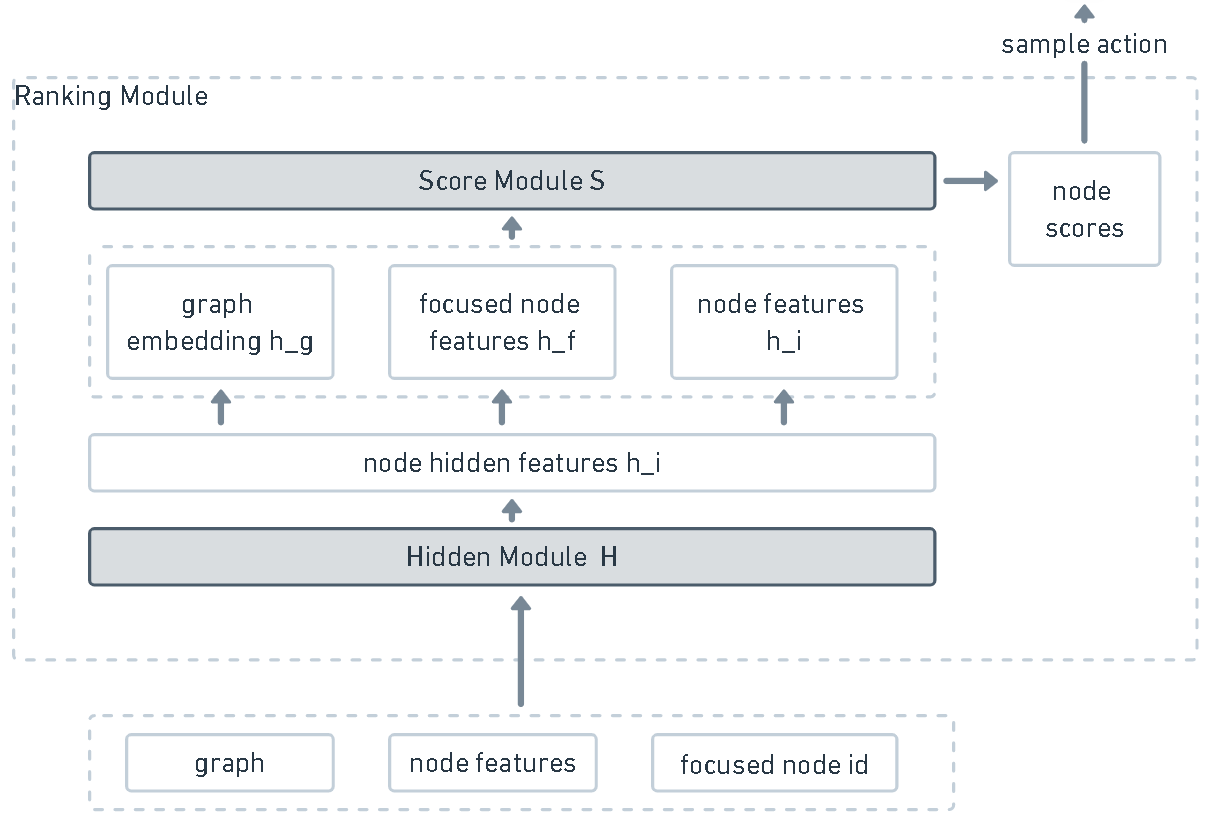}
\caption{\textbf{Ranking Module}. It is composed of a hidden layer $H$ that can be a MLP or a GAT, and a score module $S$ that is a single-layer MLP. $h_i$ stands for the hidden features corresponding to node i, $h_f$ is the same thing where $f$ is the id of the focused node. $g_f$ correspond to the graph hidden features, obtained by aggregating all node hidden features. Node scores are then used to form a distribution from which the action is sampled.}
\label{fig:ranking_module}
\end{figure}

Our recommender policy consists of a ranking module that scores each node in the graph. Scores generate a probability distribution over nodes from which actions are sampled. During training a Multinomial distribution is used, while in evaluation, the action is simply the \textit{argmax} of scores.

The Ranking Module includes a hidden module $H$ that produces node embeddings, and a score module $S$ which calculates scores for each node. $H$ can be either a MLP (Multi layer perceptron) or a GAT (Graph Attention Network). MLP is faster to train but considers only individual node features, while GAT is more expressive but slower to train, relying on message passing and learning to weigh the contributions of each node's neighbor. 

We picked GAT over GCN because GCN did not pass our preliminary experiments and GAT did. We were unable to parameterize a GCN that could learn a simple function of node features in a Supervised Learning setting. We tried GCN in the following parameter space: {$\#GConv \space layers \in \{1, 2, 3\}$ ;  $layer size \in \{ 16, 32, 64\}$; aggregation function $agg \in\{sum, mean, max \}$ ; learning rate $lr \in \{0.01, 0.001, 0.0001\}$; dropout values $drop \in \{0, 0.5\}$.

We hypothesize this was due to the labels for our supervised task depending solely on the nodes own features. GCN would construct node representations by indiscriminately aggregating their neighborhoods, confounding neighbor features with their own. By learning to weigh edges, GAT could selectively focus on the node's self-edge. For this reason, and given the importance of a node's own features to our task, we chose to use GAT in our GNN policy.

\subsubsection{Modules}

A brief explanation of the sub-modules of our Ranking Module.
 
\textbf{Input.} Input to the ranking module consists of an adjacency matrix $A$, the focused node id $n_f$, node features $X$ (strategy $s_i \in {0,1}$, and normalized node degree), and an action mask that disables invalid actions. (the node itself or nodes that are already neighbors)

\bigskip

\textbf{Hidden Module.} The hidden module computes node hidden features $h_i = H(x_i)$. It can be a MLP or GAT, we study both.

\bigskip

\textbf{Score Module.} The score module is a single-layer MLP that takes the hidden features $h_i$ of each node, concatenated with $h_f$, those of the focused node, and graph features $h_g$ obtained by summing over all node features. $score_i = S(h_g, h_f, h_i)$, \qquad $h_g = \sum_{i \in V} h_i$

\subsection{Training}

At each step, the Ranking Module ranks nodes based on their own features, the focused node's features, and whole-graph features (obtained by aggregating all node features). The MLP hidden layer  is unable to use neighborhood information and so it will rely solely on the relationship between features of nodes and graph features. The GAT hidden layer uses neighborhood information and learns to weigh the contributions of each neighbor to a node's hidden features.

Each of the learning curves plotted in this section is the average of the 3 best runs for that configuration, while the clouds around them delimit maximum and minimum values. In plots for \#cooperators, the reward value corresponds to $rew = 2 * (coops - 0.5)$, such that convergence to 1.0 cooperators give 1 reward and to 0.0 cooperators gives -1 reward. One training step corresponds to one batch of environment time-steps, or rewire updates. 

\textit{Mean Action Strat} plots the mean strategy of recommended nodes over training (when $D=0$ and $C=1$), whereas \textit{Mean Action Degree} tracks the mean degree of recommended nodes.

A difference in performance between training and evaluation may be observed and can be explained by the fact that in evaluation actions always results of picking the highest score, whereas during training, they're sampled from a distribution to ensure exploration.


\begin{table}[htbp]
\centering
\caption{Recommender reward functions being studied.}\label{tab:rewards}
\resizebox{1\linewidth}{!}{
\begin{tabularx}{1.4\linewidth}{llX}
\toprule
\textbf{Short} &  \textbf{Recommender Name} &  \textbf{Description}\\
\midrule
\textbf{\#rewires} & Engagement maximizer & Total amount of rewire requests made to the recommender in each episode - as opposed to total number of time-steps (strategy updates + rewire updates).\\
\textbf{\#coops} & Aligned recommender & Final number of cooperators. A natural metric for social good in our toy environment.\\
\bottomrule
\end{tabularx}
}
\end{table}
        
\subsubsection{N=10}
\begin{figure}[htbp]
	\centering
	\subfigure[\#coops.]{\label{fig:n10_coop_metric} 		\includegraphics[width=0.31\linewidth]{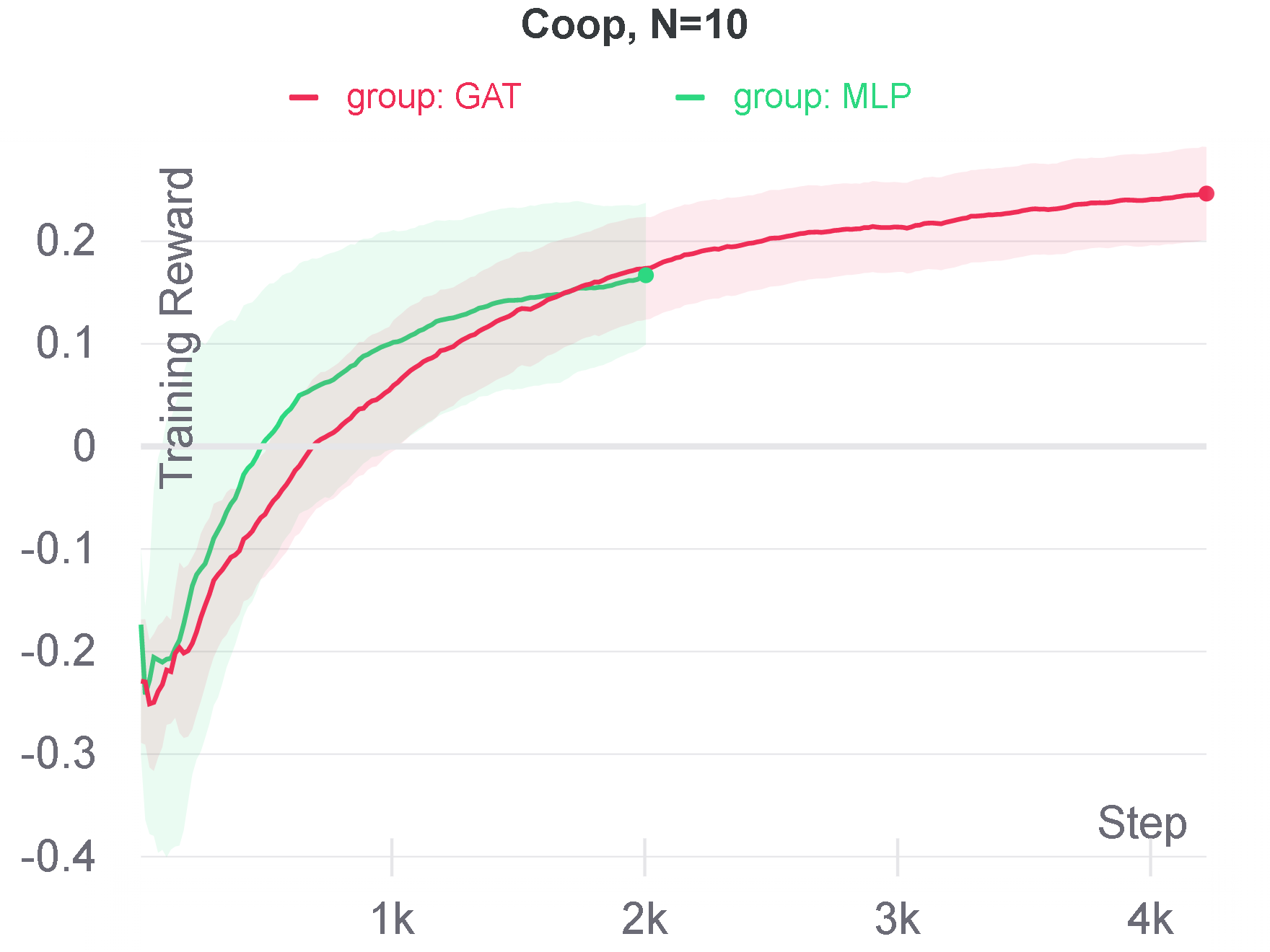}}
	\subfigure[Strategy.]{\label{fig:n10_coop_strat}
	\includegraphics[width=0.31\linewidth]{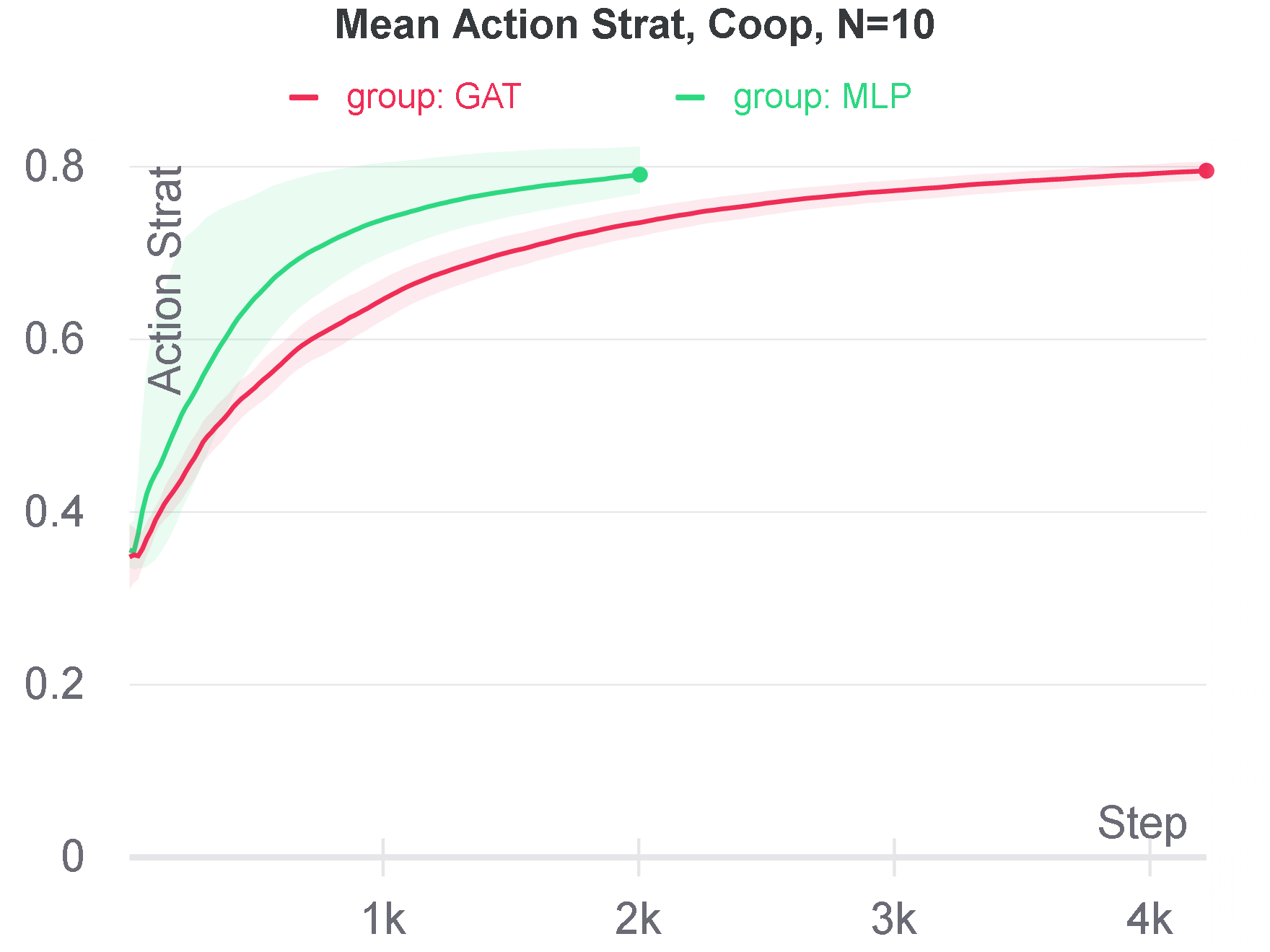}}
	\subfigure[Degree.]{\label{fig:n10_coop_deg} 		\includegraphics[width=0.31\linewidth]{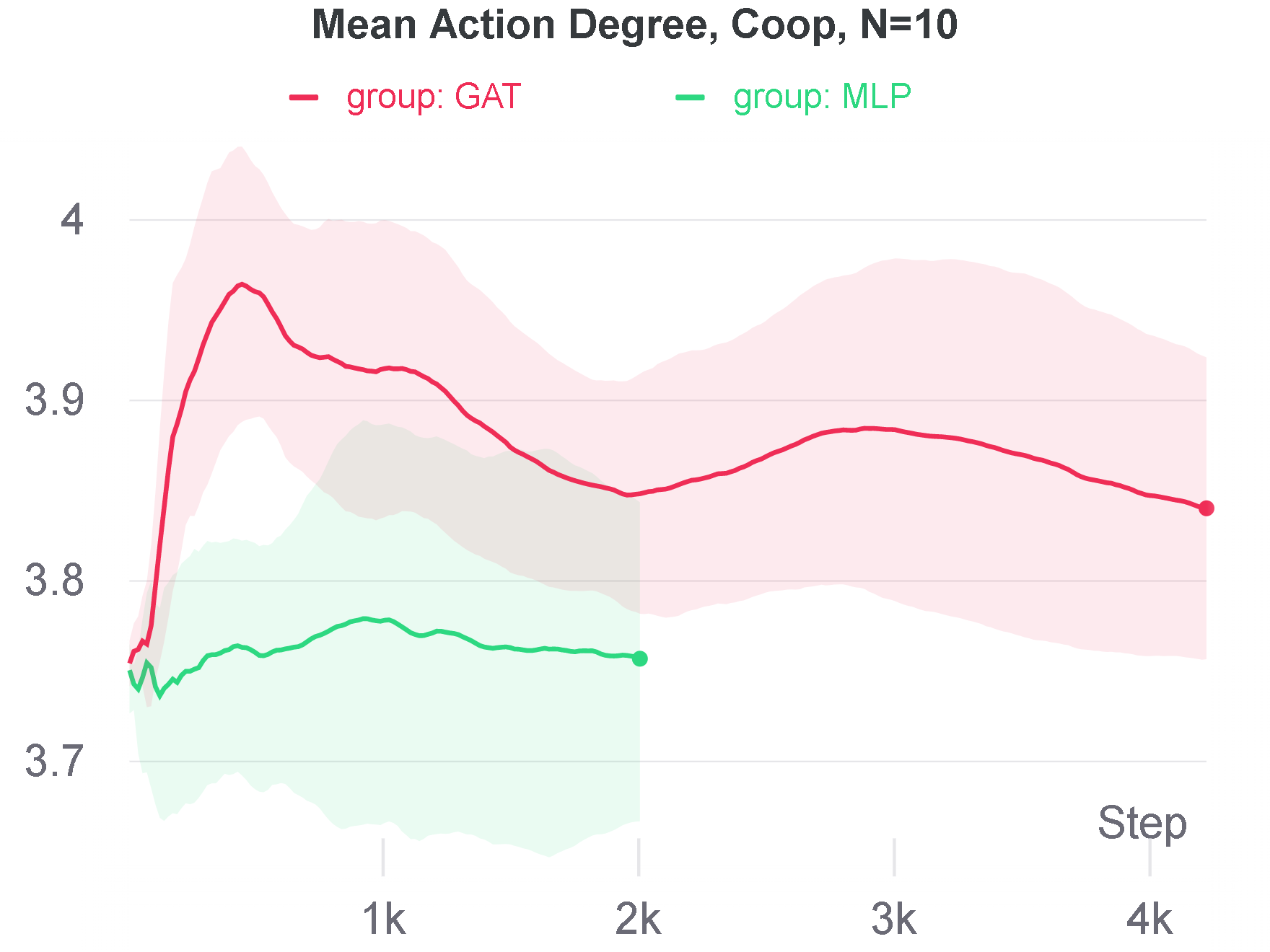}}
	\caption{Learning curves for \textbf{\#cooperators} at $N=10$. We can see both policies learn and GAT surpasses MLP. Both policies also beat heuristic baselines. We can see GAT consistently recommends slightly higher degree nodes and eventually reaches MLP's average action strategy.}
	\label{fig:learning_coop_small}
\end{figure}

\begin{figure}[htbp]
	\centering
	\subfigure[\#rewires.]{\label{fig:n10_rewire_metric} 		\includegraphics[width=0.31\linewidth]{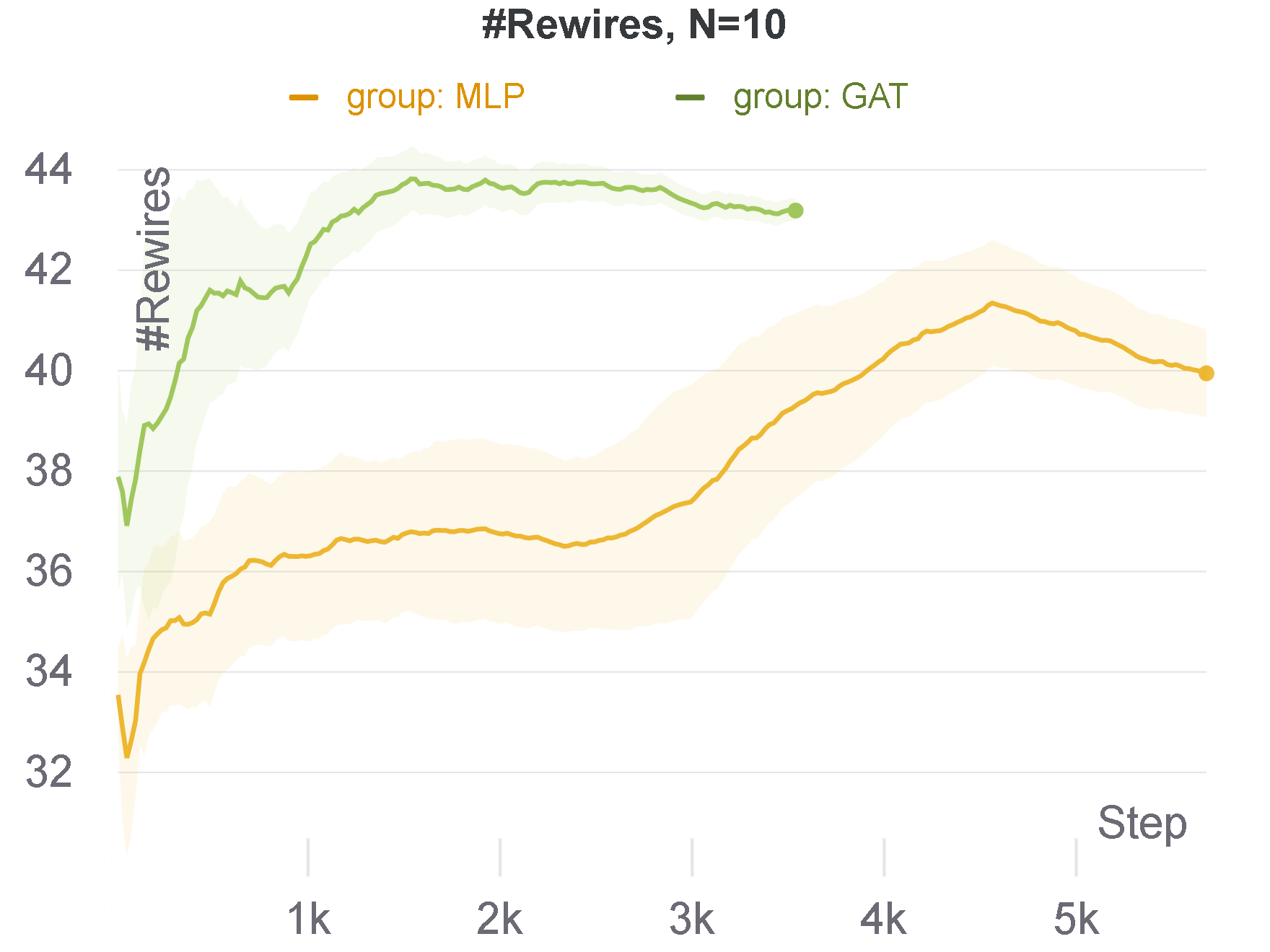}}
	\subfigure[Strategy.]{\label{fig:n10_rewire_strat}
	\includegraphics[width=0.31\linewidth]{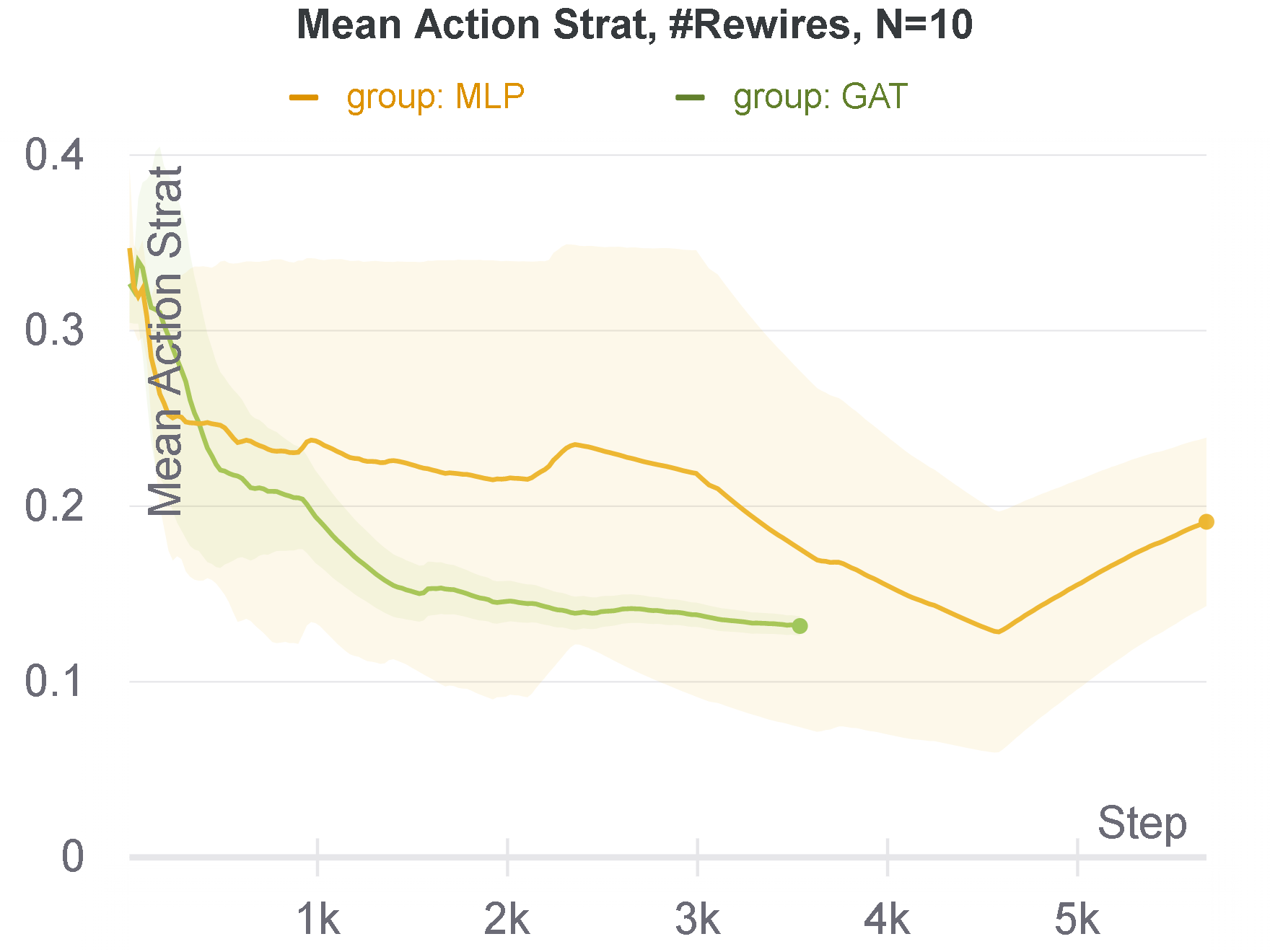}}
	\subfigure[Degree.]{\label{fig:n10_rewire_deg} 		\includegraphics[width=0.31\linewidth]{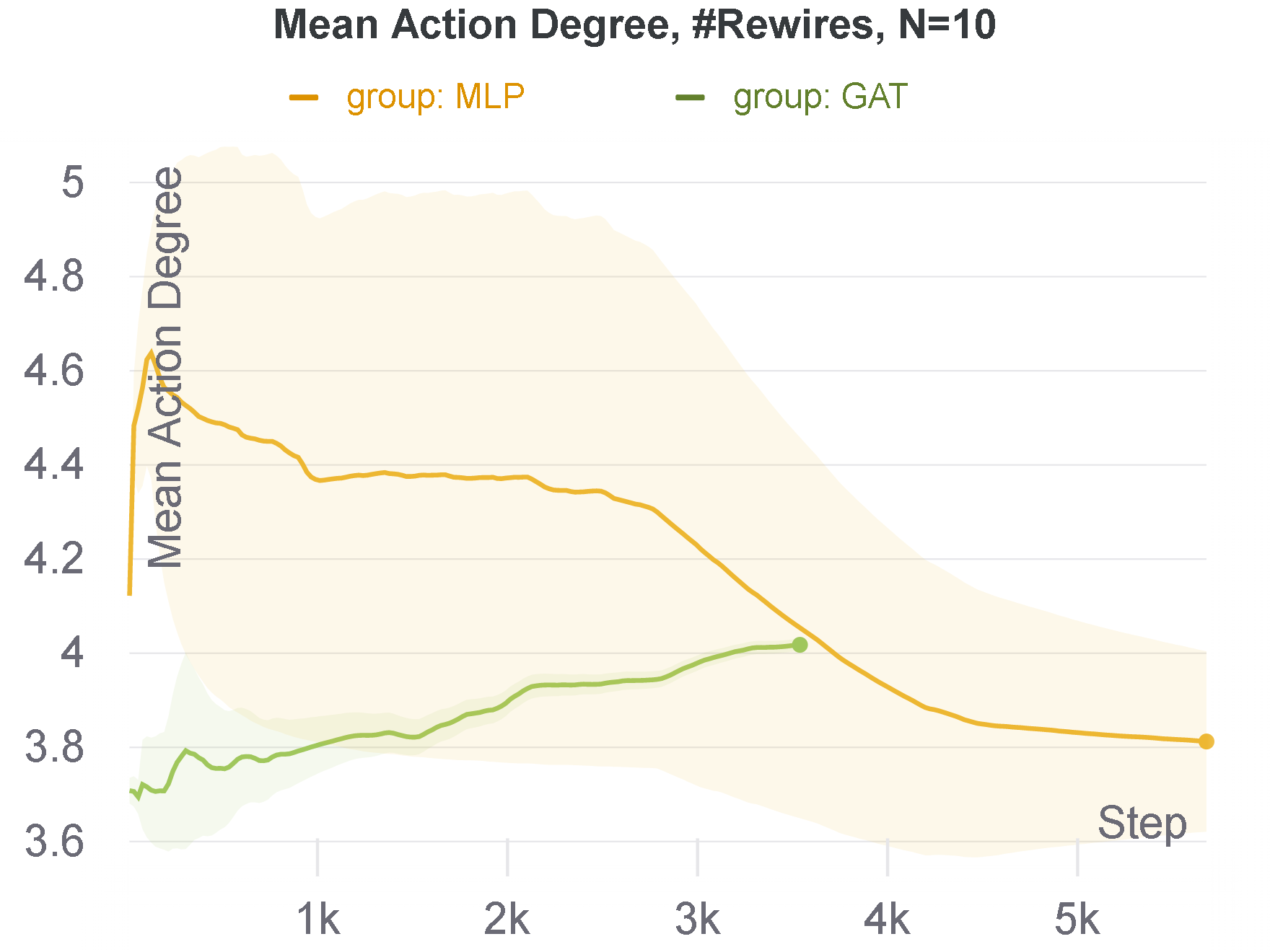}}
	\caption{Learning curves for \textbf{\#rewires} at $N=10$. We can see both policies learn and GAT is consistently better than MLP. In this case, only GAT beats all heuristic baselines. In this case \#rewires seems to correlate with lower average action strategy (rapid early increase in score) and lower action degree (decrease in score coinciding with increasing action degree).}
	\label{fig:learning_rewire_small}
\end{figure}


The environment converges quickly when $N=10$. We are able to learn policies that do better than any heuristic for either reward function. The modest improvements in \#coops may be due to the size and convergence speed of the environment leaving little room to act strategically. 

GAT policy performs slightly better than MLP on \#coop but considerably better on \#rewires. The \#coop policy trained for $N=30$ actually performed better on this environment than any of the $N=10$ policies we trained. We observe the behavior of learned policies usually resemble the best heuristics in most metrics. However, there are noticeable differences in action strategy and action degrees that might account for learned improved behavior.

\subsubsection{N=30}

\begin{figure}[htbp]
	\centering
	\subfigure[\#coops.]{\label{fig:n30_coop_metric} 		\includegraphics[width=0.30\linewidth]{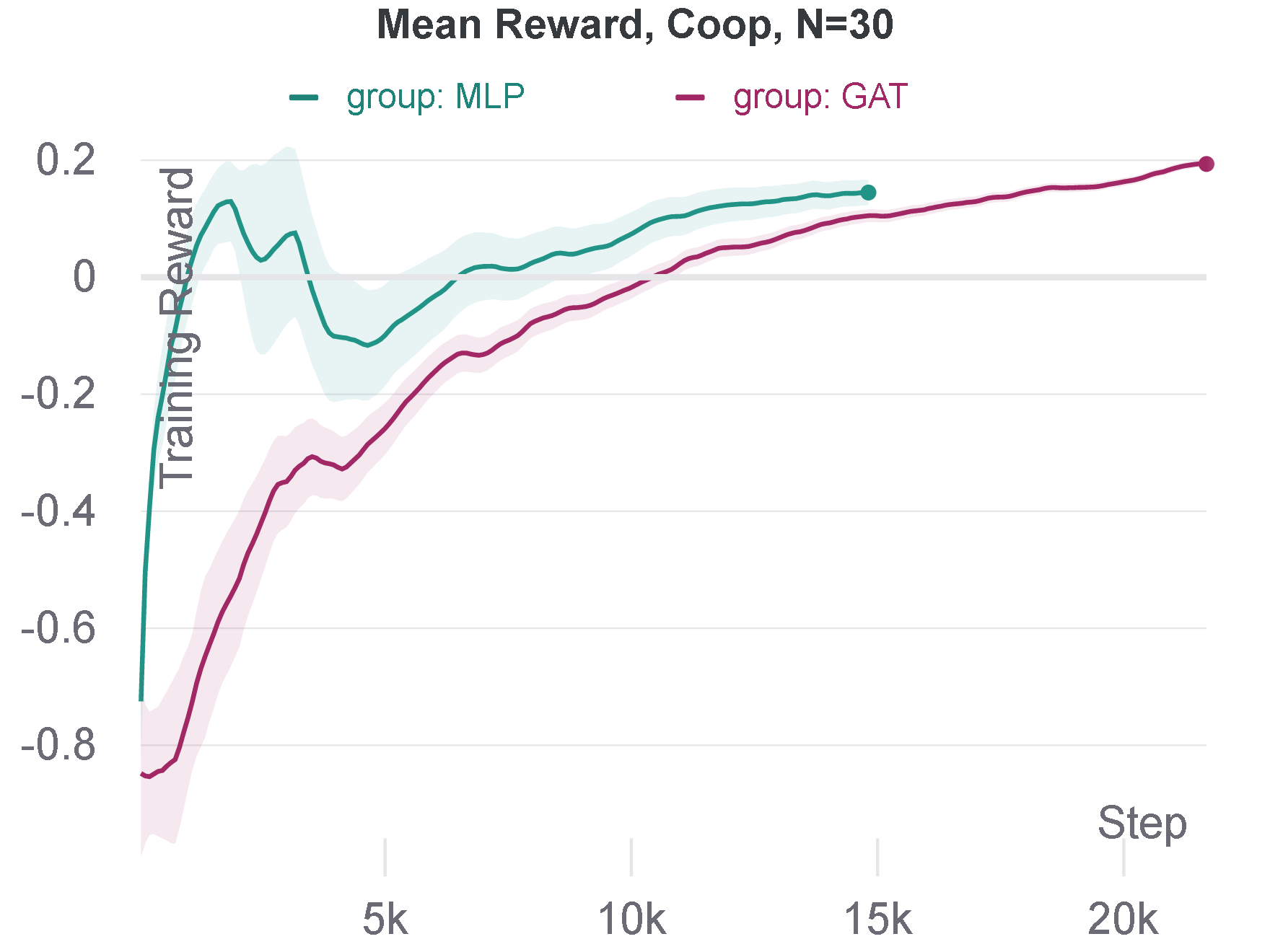}}
	\subfigure[Strategy.]{\label{fig:n30_coop_strat}
	\includegraphics[width=0.30\linewidth]{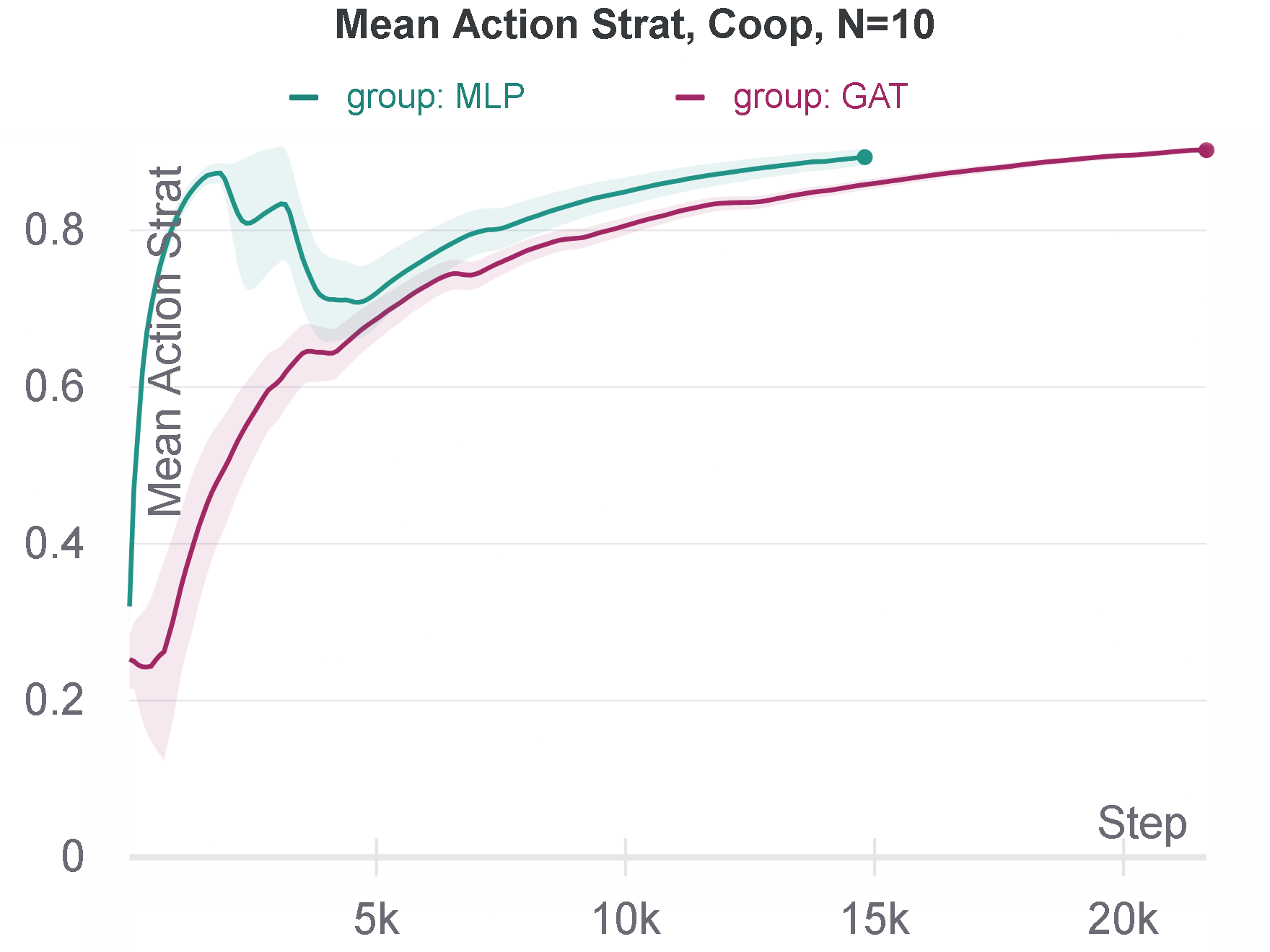}}
	\subfigure[Degree.]{\label{fig:n30_coop_deg} 		\includegraphics[width=0.30\linewidth]{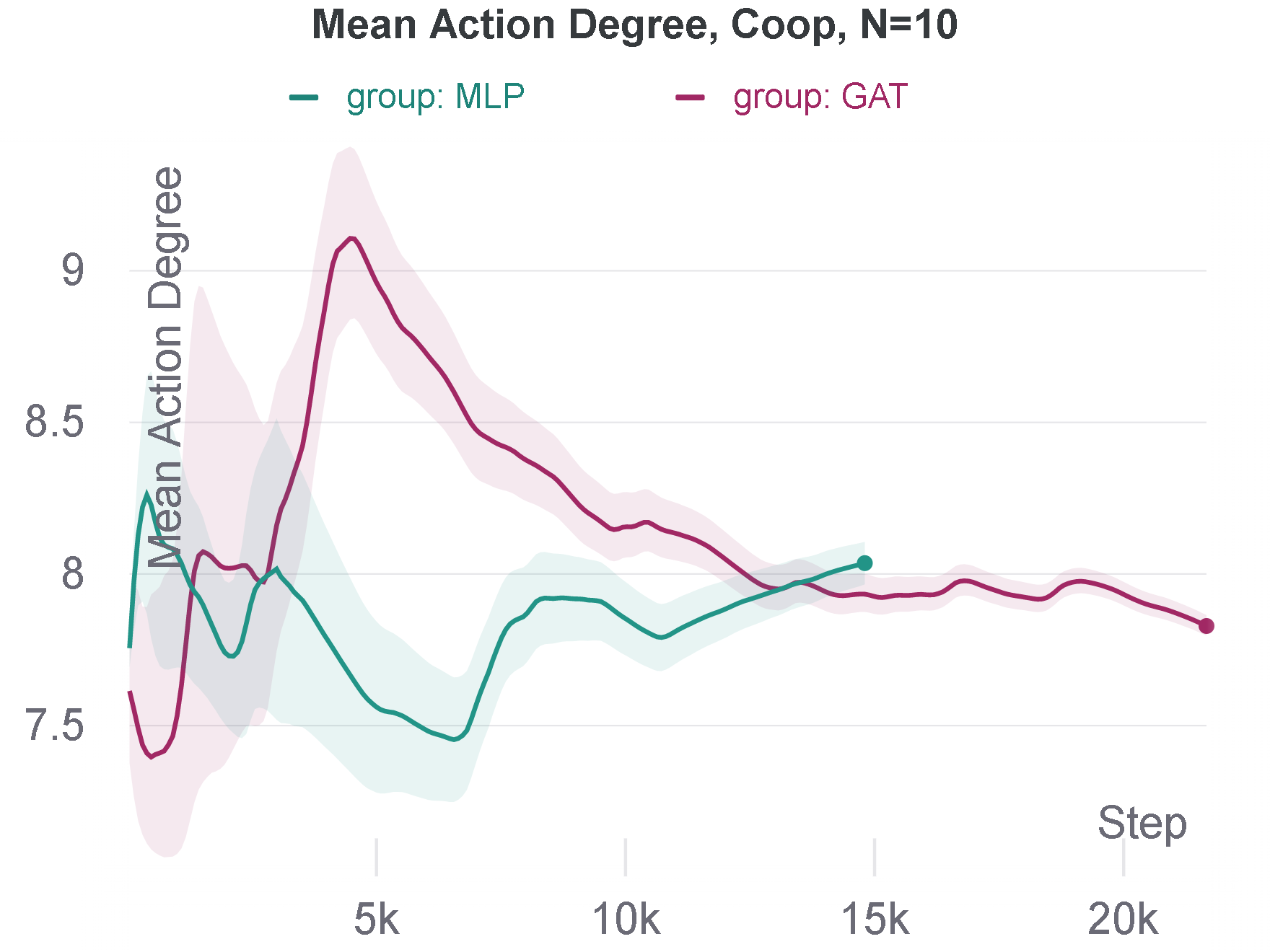}}
	\caption{Learning curves for \textbf{\#cooperators} at $N=30$. Similarly to N=10, we can see both policies learn and GAT slightly surpass MLP. Both policies also beat heuristic baselines. Despite noticeable variance in mean action degree, mean action strategy seems to be the determining factor in this case.}
	\label{fig:learning_coop_balanced}
\end{figure}

At $N=30$, MLP policies significantly outperformed GAT ones both for \#coop and \#rewires. GAT's higher complexity might require more training episodes than MLP, which could explain its lower performance, as both policies were trained using the same resources. 
\subsubsection{N=100}

\begin{figure}[htbp]
	\centering
	\subfigure[\#coops.]{\label{fig:n100_coop_metric} 		\includegraphics[width=0.31\linewidth]{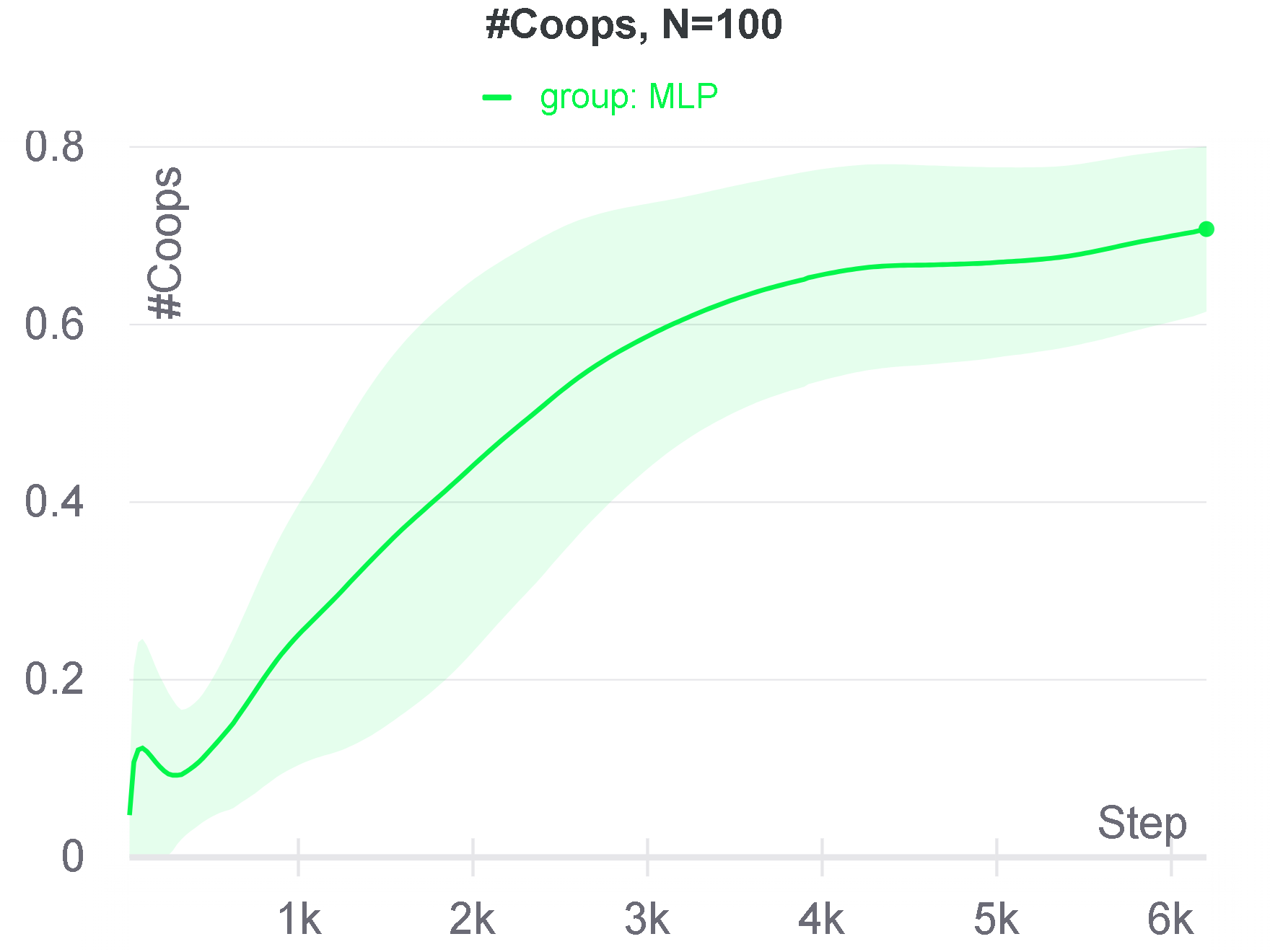}}
	\subfigure[Strategy.]{\label{fig:n100_coop_strat}
	\includegraphics[width=0.31\linewidth]{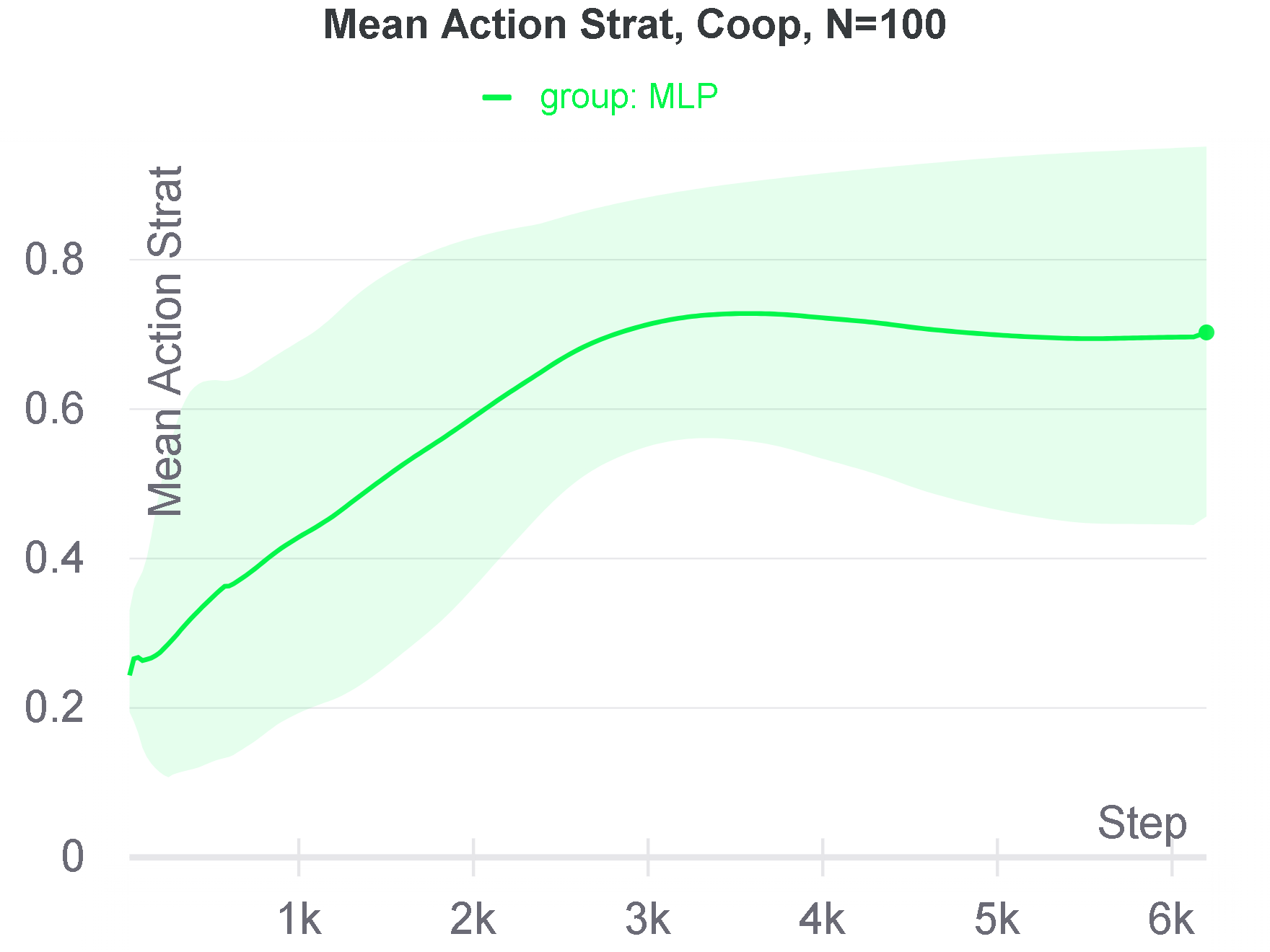}}
	\subfigure[Degree.]{\label{fig:n100_coop_deg} 		\includegraphics[width=0.31\linewidth]{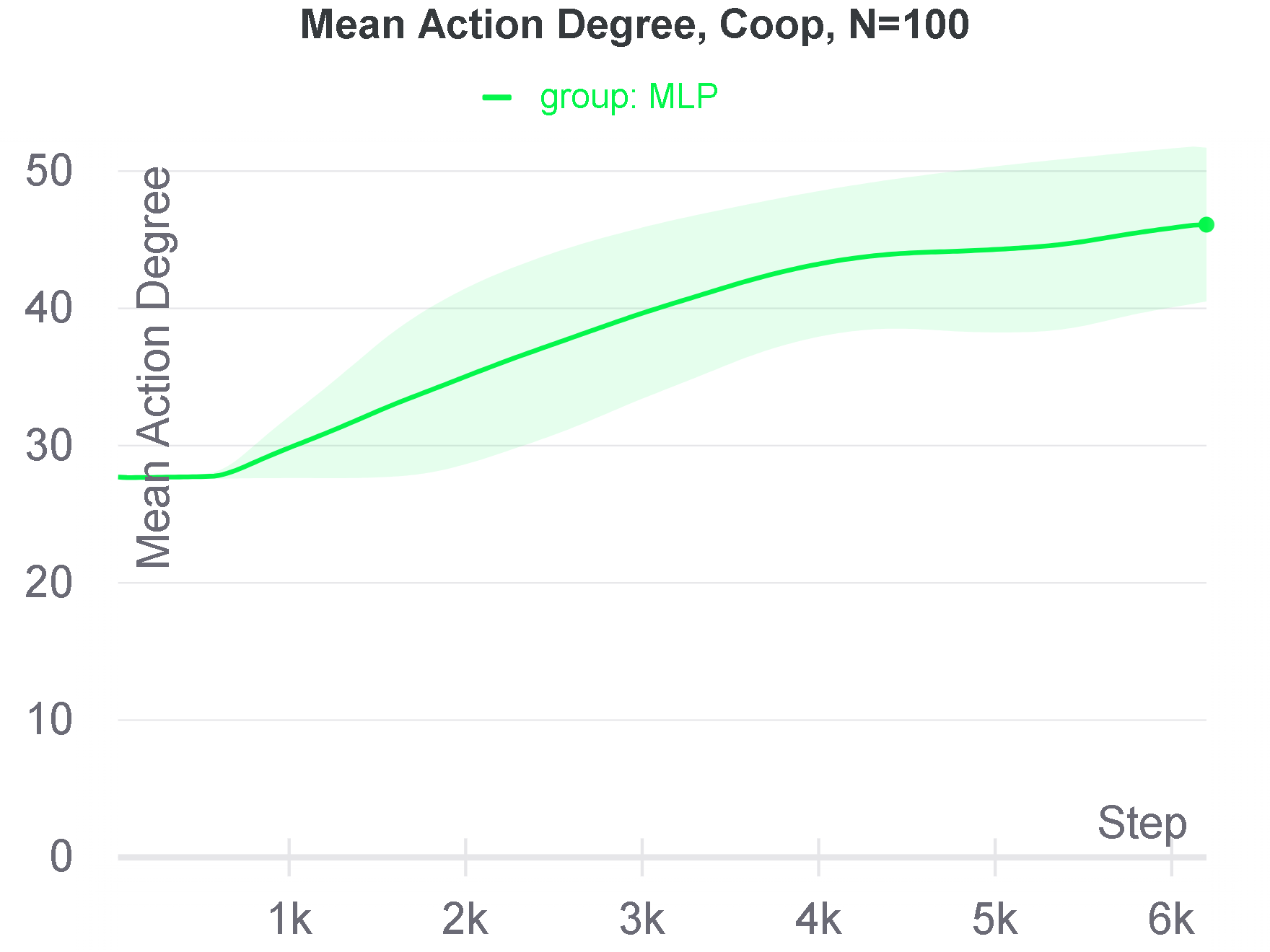}}
	\caption{Learning curves for \textbf{\#cooperators} at $N=100$. We weren't able to train a GAT policy in this environment. GAT is computationally more expensive to train than MLP, and at this size, computational constraints start limiting our architecture.}
	\label{fig:learning_mlp_coop_med}
\end{figure}

At $N=100$, MLP \#coop matches the top heuristic's performance for \#coops. Our MLP trained to maximize \#rewires on N=100 does not beat to top baselines, but the GAT policy trained on $N=10$ is able to beat all heuristics by a large margin.

We have not been able to make GAT policies converge on environments where $N=100$ or $N=500$. Additionally, \#coop  seems to converge much more smoothly than \#rewires at every value of N. This may be explained by the fact that there is a mapping between observations and \#coops (node strategy features), whereas \#rewires is merely counted and given as reward at the end of an episode, without markers in the observable state. We haven't tested this hypothesis.

\subsubsection{N=500}
At $N=500$, despite having trained policies specifically on $N=500$, the best performing policy is the one trained for MLP \#Coop $N=100$ and it beats all heuristics. Again, we hypothesize that policies in smaller environments converge faster, and that difference makes up for differences of dynamics in networks of $N=100$ and $N=500$.

\begin{figure}[h]
\centering
\includegraphics[width=1\linewidth]{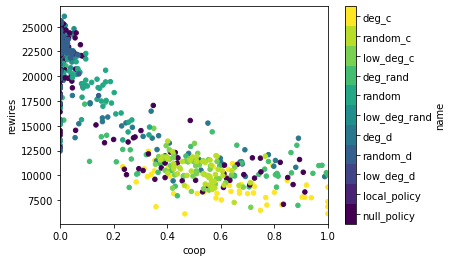}
\caption{Scatter plot of runs for different recommender heuristics. The axes are \textbf{\#coops} and \textbf{\#rewires} for $N=500$. We can observe a sort of convex Pareto frontier and a rough ordering among heuristics, with policies that recommend cooperators leading to higher \#coop and lower \#rewires, and the opposite for policies that recommend defectors.}
\label{fig:scatter_big}
\end{figure}

To conclude this section, we show that \textbf{it is possible to learn to control evolutionary network dynamics by means of recommendations}. It is even possible to learn policies that outperform all heuristic combinations of strategy and node degree for some configurations of our environment. We expect that the configurations where we were unable to learn policies might be solvable with more fine-tuning, or simply applying more resources.

\subsection{Analysis}

We began this toy experiment by asking "\textit{How can we evaluate the consequences of using user retention as a reward function?}". More generally, to explore how we could evaluate and compare reward functions with respect to the societal effects caused by recommenders trained to optimize them. So we defined a toy environment to represent society and used reinforcement learning to optimize these reward functions.

Having evaluated our learned policies as well as many heuristics, we can ask the question in the context of our toy model: Do engagement maximizers perform significantly worse than aligned recommenders? The answer is, perhaps unsurprisingly, yes. In all our settings, engagement optimizers will lead the environment to low final numbers of cooperators. The only exception was the environment where $N=30$, where the best performing engagement policy produces a final fraction of cooperators around 0.5 which is high in comparison with other heuristics.  

This suggests something we already expected. With respect to our toy environment, misaligned recommenders like the engagement maximizer may or may not lead to bad outcomes for society, but they are worse than aligned recommenders. In the case of $N=500$ \cref{fig:scatter_big}, these two reward functions are at odds with one another, at least for our heuristics which trace a convex curve in the axes of \#coop and \#rewires.

Strategies that recommend cooperators generally lead to higher convergence to full cooperation than the others, we don't observe such a pattern in policies that maximize \#rewires.

With regards to topology, it is very clear that heterogeneity increases with mean action degree and that it is generally lower for heuristics that recommend cooperators than for the others. 

Aligned policies also tend to lead to more heterogeneous networks than misaligned ones, again with the exception of $N=30$.

One limitation in these experiments is that we are considering worlds where there is only one rewiring policy and simulating what happens if people keep using it. In the real world, if a recommender is bad enough, people will just not use it at all. 

A more realistic setup would look like competition between the recommender and the null or local heuristics. In that case, there would be a bound on how bad the RS could be before users decide they might as well not rewire social ties at all, or only within their local environment, respectively. 

If learning happens in these conditions, we might see a "bait-and switch" strategy emerge. The recommender might begin by being advantageous to users while trying to get adopted by as many users as possible, and switching to an "exploit" dynamic once dominant in the population, to further maximize its reward function.

\section{Experiment: Competing RS to Address the Alignment Problem}
\label{sec:competition}
\subsection{Competition Dynamics}

We extend our environment to allow competition between recommenders. This is achieved by attributing a recommender strategy to each node and allowing them to evolve in the same way that game strategies evolve. That is:

\paragraph{Recommender Update} At each time-step, there is a chance that the update performed is a recommender update. That is, the user may change which recommender she is using to rewire ties. In that case, an agent selects a random neighbor and imitates its rewiring strategy with probability p weighed on their fitness difference given by the Fermi update. (\cref{eqn:fermi}) 

Recommenders are exclusive, meaning they will only recommend nodes from among their own users. Keep in mind users still interact with their neighbors, even if they are using other recommenders. This reflects what we see in the world, where recommender systems (mostly) only have information about their own users.

We introduce a second time-scale ratio $W2 = t_m / (t_e + t_a)$ to regulate the relative frequencies of mediator updates ($t_m$) and two other kinds of updates. (strategy $t_a$ or structural $t_e$) 

All competition runs are initialized with 1000 nodes and a time-limit of $10^5$ time-steps. The effects of competition were not as clear on populations of 500. 
Recommender updates use a different temperature parameter $beta_{med} = beta * 10 = 0.05$ because we observed the impact of mediator updates was negligible using 0.005. 
Given the stochastic nature of our simulations, all presented results are averaged over 30 runs.

\begin{figure}[h]
\centering
\includegraphics[width=1\linewidth]{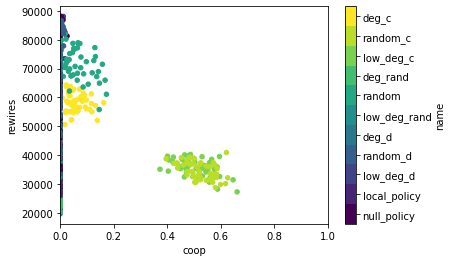}
\caption{Scatter plot where the axes are \#coops and \#rewires for N=1000. We can observe a sort of Pareto frontier and a rough ordering among heuristics, with policies that recommend cooperators leading to higher \#coop and lower \#rewires, and the opposite for policies that recommend defectors.}
\label{fig:scatter_comp}
\end{figure}

\subsection{Competing Baselines}

We begin by taking an environment dominated by the local heuristic and introducing recommenders into $10\%$ of the population. 90\% of nodes use the local heuristic NO MED, while each node of the remaining 10\% has one of the others. (GOOD, BAD, RANDOM, FAIR) 

We began with measuring the effects of competition over various timescale combinations W and W2. ($\{0.5, 1, 2, 3, \infty\}$ and $\{0.01, 0.03, 0.1, 0.5, \infty\}$ respectively)

We observed that for $W1=\inf$ (no strategy updates, only rewires and mediator updates), the initial conditions remain practically the same, while for $W2=0$ the initial conditions remain the same due to there being no mediator updates.

\begin{figure}[htbp]
	\centering
	\subfigure[W1=1. GOOD and NO MED share dominance over the others.]{\label{fig:w2_sweep_med_freqs_w1_1-0} 		\includegraphics[width=0.48\linewidth]{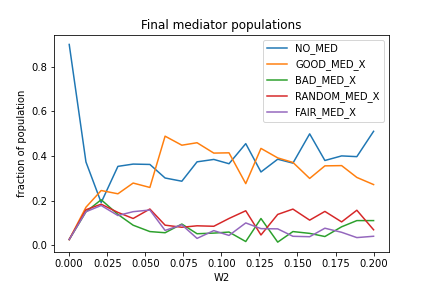}}
	\subfigure[W1=1.2. Same as W1=1 but less pronounced]{\label{fig:w2_sweep_med_freqs_w1_1-2} 		\includegraphics[width=0.48\linewidth]{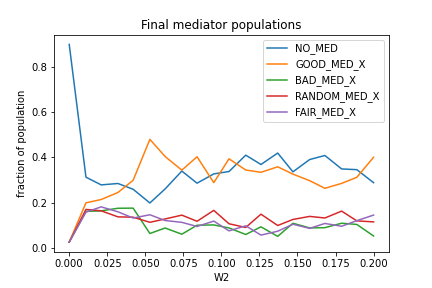}}
	\subfigure[W1=1.6. Even less pronounced than W1=1.2 .]{\label{fig:w2_sweep_med_freqs_w1_1-6} 		\includegraphics[width=0.48\linewidth]{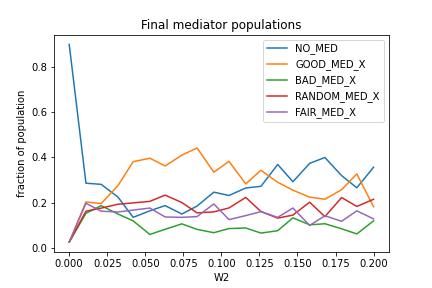}}
	\subfigure[W1=2. Tighter than W1=1 but BAD is at the bottom, while GOOD and NO MED share the top by a smaller margin. ]{\label{fig:w2_sweep_med_freqs_w1_2-0}
	\includegraphics[width=0.48\linewidth]{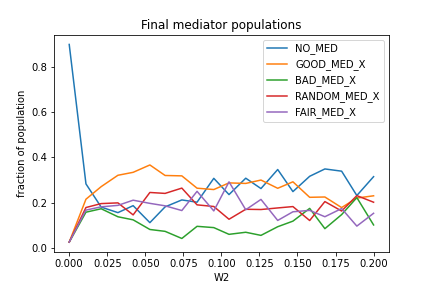}}
	\caption{Final frequencies of mediators as functions of W2. }
	\label{fig:w2_sweep_med_freqs}
\end{figure}

We find a critical region where NO MED does not have a majority around $W1 \in [1,2]$ and $W2 \in [0.03,0.1]$. 
We investigate this range more closely in \cref{fig:w2_sweep_med_freqs} and observe that GOOD and NO MED have a tendency to dominate over the others. That is most pronounced at $W1=1$. As W1 increases, the gap between strategy populations becomes less evident, although BAD takes the bottom place more convincingly.

\subsection{Adoption Experiments}

\textbf{Questions for adoption experiments:}
\begin{enumerate}
\item Will engagement maximizers be adopted and thrive in a society of local heuristics?  What about aligned recommenders?
\item In a society dominated by engagement maximizers, will an aligned recommender be adopted and thrive?
\item Do worlds were RS competition exists lead to better outcomes for society than worlds dominated by engagement maximizers?
\end{enumerate}


From our experiments, competition results are clearest in environments where N=1000. Due to computational and time constraints, we were unable to learn policies directly on these large environments. Despite this, we pick the heuristics that score highest in our metrics of \#coops and \#rewires to use as proxies for policies trained explicitly to maximize them. We pick "Random cooperators" as our aligned recommender and "Random" as our engament maximizer.

We run one trial for each recommender in environments with local-heuristic majorities (90\% local). We want to know whether recommenders maximizing user-engagement would be able to take over a population starting from a small seed. We ask the same question for recommenders maximizing cooperation.

\begin{figure}[htbp]
	\centering\
    \subfigure[90\% local heuristic vs. 10\% minority engagement maximizer.]{\label{fig:comp_local_v_rand} 		\includegraphics[width=0.44\linewidth]{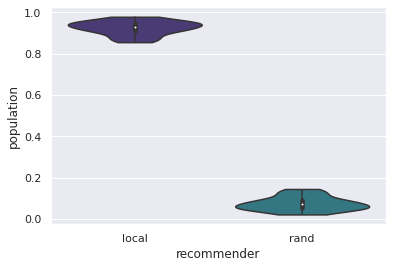}}
	\subfigure[90\% local heuristic vs. \%10 aligned recommender.]{\label{fig:comp_local_v_rand_c} 		\includegraphics[width=0.44\linewidth]{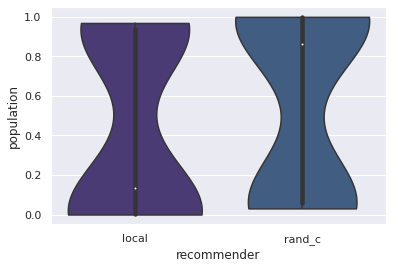}}
	\subfigure[90\% engagement maximizer vs. 10\% minority aligned recommender.]{\label{fig:comp_rand_v_rand_c} 		\includegraphics[width=0.44\linewidth]{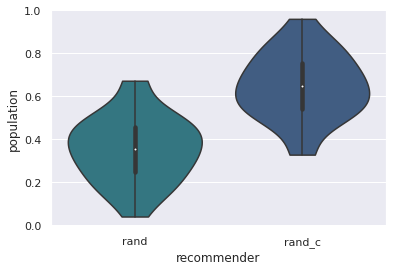}} 
    \caption{Distribution of final populations in competition. Engagement maximizer doesn't manage to get adopted, while aligned recommender gets adopted slightly above 50\% of the population. Aligned recommender is able to dominate in an environment initially dominated by the engagement maximizers. Initial conditions in the captions of sub-figures.}
	\label{fig:comp_violins}
\end{figure}

We can observe that a diverse competition scenario (Local vs Many) leads to high cooperation and wide adoption of recommenders in our toy environment. This is consistent with the suggestion that competition between RS is a desirable feature to implement in the world, protecting it from misaligned recommenders.

\begin{table}[!ht]
    \centering
    \caption{Metrics resulting from competition between optimizer policies and local heuristic. Averages over 30 runs. Both are adopted over the local heuristic. "Final prop. of start maj." stands for the average final proportion of the policies that began as the majority (local, local, and engagement respectively in the competition scenarios below).}
    \resizebox{1\linewidth}{!}{\begin{tabular}{lccc}
    \toprule
        {Heuristic} & {Coops} & {Rewire} & {Final prop. of start maj.} \\ \midrule
        Engagement & 0.06372 &  71487.22 & - \\
        Engagement vs Local majority & 0.0 & 39301.033 & 0.925 \\
        Local & 0.00000 & 33934.28 & - \\ 
        Local vs Many & 0.432 & \bfseries 277427.633 & 0.227 \\
        Aligned vs Local majority & 0.362 & 225244.533 & 0.427 \\ 
        Aligned vs Engagement majority & 0.469 & 417636.18 & 0.443 \\ 
        Aligned & \bfseries 0.52152 & 34542.48 & - \\ 
    \bottomrule
    \end{tabular}}
    \label{baseline_comp}
\end{table}

\section{Previous Work}
\label{sec:related}
\subsection{Societal Impacts of RS}
There have been qualitative inquiries into the societal impacts of recommender systems, on which we based our modeling interface: RS as multi-stakeholder environments \cite{milano_recommender_2020}. A taxonomy of ethical issues associated with RS \cite{milano_recommender_2020}. An overview of consequences of widespread RS \cite{zoetekouw_critical_2019}. A taxonomy of interactions between humans and intelligent software agents \cite{burr_analysis_2018-1}. 

RS alignment has been connected to the AI alignment problem \cite{stray_what_2021} along with illustration of common patterns in current RS design. The same work also offers an overview to "higher-level approaches" to RS alignment: 1) Better metrics; 2) Participatory recommenders; 3) Interactive value learning; 4) design around retrospective judgement. However, they do not mention the role of competition among RS.

\subsection{Rewiring in Dilemmas of Cooperation}

Work in evolutionary game theory has seen setups where networked agents play social dilemmas with rewiring of social ties \cite{santos_cooperation_2006}. 
Recommendation mechanisms in spatial public goods games have also been modeled before, although the recommendations were made by other agents instead of by a central mediator. \cite{yang_role_2013}

\subsection{Learning to Control Graph Dynamics}

Graph neural networks have been used in conjunction with deep reinforcement learning to solve graph optimization problems. \cite{darvariu_goal-directed_2021} 
We base our training architecture on recent work on the control of dynamical processes in graphs through node-level interventions, \cite{meirom_controlling_2021} and train it with Proximal Policy Optimization (PPO) \cite{schulman_proximal_2017} to obtain rewiring strategies that optimize different metrics in our model.

To the best of our ability, we couldn't find previous work using RL to solve a task of mediation in evolutionary game theory. The closest application we've been able to find was RL being used to learn policies of partner selection for individual agents in the iterated prisoner's dilemma. \cite{anastassacos_partner_2020}

\subsection{RS competition}

Strategic dynamics in content production and consumption may lead to the failure of classical principles of RS in maximizing social welfare. The need to avoid such a failure by revisiting those principles with game theory and multiple stakeholders in mind has inspired a whole research agenda. \cite{kurland_rethinking_2019} 

In the same game theoretic paradigm, competing recommenders have been studied \cite{izsak_search_2014} - although not with respect to externalities. Previous work has approached the cost that competition between strategic mediators would impose on the population of agents they mediate \cite{babaioff_mechanism_2015},  although not in the context of recommendation systems.

\section{Conclusion}
\label{sec:conclusion}

This work had a broad scope in order to shed light on how an end-to-end study of reward functions for recommender systems might be done. Recommender alignment is a pressing and important problem. Attempted solutions are sure to have far-reaching impacts. The least we could do is develop methods for evaluating and comparing tentative solutions with respect to those impacts. We synthesized a simple abstract modelling framework to guide future work.

Namely, a model must include an underlying environment, users with partial information, and RS with global information; notions of utility (derived from the environment) for individual users, society, and the RS; and a notion of recommendation.

A toy experiment was run to show our framework in action. 

1. We wanted to know whether user retention is a good idea as a proxy for social good. We adapted an existing model to implement our modelling interface and ran several simulations to see how it behaves under different heuristic rewiring strategies. 

An ordering of heuristics is found with respect to speed of convergence to full cooperation. 
Surprisingly, recommending cooperators to cooperators and defectors to defectors leads to worse outcomes than a random policy.

2. Then we used proximal policy optimization and graph neural networks to learn to control graph dynamics through recommendations. We did this just to obtain policies explicitly optimized for the reward functions we wanted to compare.

We verify it is possible to learn to control graph dynamics using recommendations. Analyzing environment metrics, we conclude that engagement maximizers generally lead to worse outcomes than aligned recommenders but not always.

3. After that we extended our toy environment to allow for recommender competition within the same population and compared simulations of it under competition with simulations under a single recommender. 

We find that recommender competition generally makes our society better-off. Aligned recommenders are found to be able to replace both local heuristics and engagement maximizers. The engagement maximizers we tested are not found to be able to replace local heuristics, suggesting real world RS would've needed different strategies to be adopted.

Our main focus wasn't the results of these experiments, which are simple and self-contained, but rather the fact that we were able to construct them thanks to our modelling interface, which allows for arbitrary complexity in the concrete model used. With the second interesting benefit of learning to control the evolutionary graph dynamics of our toy environment.

We hope this has laid some foundations for future work on proposals for RS alignment and their evaluation. The experience of billions of people is shaped daily by content recommendation. As society changes, so must our objective functions, therefore the need for a bottom-up adaptive system that answers to users. Aligning recommender systems is one of the critical tasks of our time.

\section{Future Work}
\label{sec:future}

The direction we're most excited about working on is in learning recommenders under competition. Especially the base case where recommenders merely need to compete with the null or the local strategies. In this case, recommenders don't learn under the assumption that they have a monopoly on their populations. Therefore, even if misaligned, they must learn to not be actively detrimental, and a little better than the baseline in order to get adopted. We expect we'll be able to find an emergent "bait-and-switch" dynamic here that is also present in the real world, where after widespread adoption, when it is more disadvantageous for users to leave because of network effects, recommenders turn to "exploit mode", caring less about user utility and more about their own.

If we were able to spend more time on learning graph dynamics, it would be edifying to apply methods of interpretability to understand what strategies the agents have learned to beat to baselines. Do they focus on converting hubs? Do they target high-betweenness nodes somehow? Are focused nodes served differently based on their strategy?

Another avenue of study would be games beyond the \textit{prisoner's dilemma}, like \textit{ultimatum}, or public goods, extending user interaction from 1-to-1 to many-to-many. Completely different environments like \textit{Sugarscape} or sequential social dilemmas \cite{leibo_multi-agent_2017} are also worth exploring.

Finally, midway through the work of this dissertation, temporal GNN libraries like Pytorch Geometric \cite{rozemberczki_pytorch_2021} became available, which would certainly make learning more efficient enabling training for longer periods and perhaps allow convergence in some of our larger environments.



\normalsize
\bibliography{references}


\end{document}